\documentclass[11pt]{article}
\usepackage{acl}

\usepackage{listings}
\usepackage{colortbl}
\usepackage{colortbl}
\usepackage[utf8]{inputenc} 
\usepackage[T1]{fontenc}    
\usepackage{hyperref}       
\usepackage{url}            
\usepackage{booktabs}       
\usepackage{amsfonts}       
\usepackage{nicefrac}       
\usepackage{microtype}      
\usepackage{caption}
\usepackage{subcaption}
\usepackage{graphicx}
\usepackage{mathtools}
\usepackage{multirow}
\usepackage{times}
\usepackage{pgfplots}
\usepackage{geometry}

\usepackage[normalem]{ulem}
\usepackage[edges]{forest}
\usepackage{array}
\usepackage{geometry}
\usepackage{wrapfig}
\usepackage{subcaption}
\usepackage{textcomp}
\usepackage{gensymb}
\usepackage{mathtools}
\usepackage{amssymb}
\usepackage{wrapfig}
\usepackage{lipsum}
\usepackage{etoolbox}
\usepackage{tikz}
\usepackage{arydshln}
\usepackage{microtype}
\usepackage{latexsym}
\usepackage{booktabs} 
\usepackage{colortbl} 

\definecolor{darkblue}{HTML}{00008B}
\definecolor{lightblue}{HTML}{ADD8E6}
\usetikzlibrary{arrows.meta}
\usepackage{ifthen}
\usepackage{xcolor-patch, xcolor}
\usepackage{color}

\definecolor{rred}{HTML}{d43e4f}
\definecolor{ppurple}{HTML}{5e4fa2}
\definecolor{ccyan}{HTML}{67c1a5}
\definecolor{ggreen}{HTML}{acdda5}
\definecolor{llime}{HTML}{e6f598}
\definecolor{yyellow}{HTML}{ffffbf}
\definecolor{llightorange}{HTML}{fdae61}
\definecolor{bblue}{HTML}{3287bd}
\definecolor{oorange}{HTML}{f46e43}
\definecolor{rrred}{HTML}{e27884}
\definecolor{pppurple}{HTML}{8f84be}
\definecolor{cccyan}{HTML}{94d3be}
\definecolor{gggreen}{HTML}{c3e6be}
\definecolor{lllime}{HTML}{edf8b6}
\definecolor{yyyellow}{HTML}{ffffd3}
\definecolor{lllightorange}{HTML}{fec590}
\definecolor{bbblue}{HTML}{a8cce2}
\definecolor{ooorange}{HTML}{f7997d}
\definecolor{hard}{HTML}{A9A9A9}
\definecolor{simple}{HTML}{D3D3D3}

\usetikzlibrary{backgrounds}
\def\donutannotate{0}
\def\donutlabelcolor{white}
\newcommand{\donutchart}[2]{
   \pgfmathsetmacro{\totalnum}{0}
   \foreach \value/\colour/\name in {#1} {
     \pgfmathparse{\value+\totalnum}
     \global\let\totalnum=\pgfmathresult
   }
  \pgfmathsetmacro{\wheelwidth}{\outerradius-\innerradius}
  \begin{scope}[rotate=90]
    \pgfmathsetmacro{\cumnum}{0}
    \foreach [count=\n] \value/\colour/\name in {#1} {
        \pgfmathsetmacro{\newcumnum}{\cumnum + \value/\totalnum*360}
        \pgfmathsetmacro{\midangle}{(\cumnum+\newcumnum)/2}
        \filldraw[draw=white,fill=\colour] (\cumnum:\outerradius) arc (\cumnum:(\newcumnum):\outerradius) --
        (\newcumnum:\innerradius) arc (\newcumnum:(\cumnum):\innerradius) -- cycle;
        \fill[darkgray!25] circle (\innerradius);
        \ifthenelse{\donutannotate=1}{
          \draw node [text=black, font=\rmfamily] (inner #2 \n) at (\midangle:{\outerradius+\wheelwidth}) {\tiny{\name}};
        }{
          \draw node [text=\donutlabelcolor, font=\rmfamily] (inner #2 \n) at (\midangle:{\innerradius+\wheelwidth/2}) {\tiny{\name}};
        }
        \node[scale=1.0, color=black, font=\rmfamily\bfseries](\innerradius) {\innername};
        \global\let\cumnum=\newcumnum
    }
  \end{scope}
  }
\makeatletter
\def\adl@drawiv#1#2#3{%
        \hskip.5\tabcolsep
        \xleaders#3{#2.5\@tempdimb #1{1}#2.5\@tempdimb}%
                #2\z@ plus1fil minus1fil\relax
        \hskip.5\tabcolsep}
\newcommand{\cdashlinelr}[1]{%
  \noalign{\vskip\aboverulesep
           \global\let\@dashdrawstore\adl@draw
           \global\let\adl@draw\adl@drawiv}
  \cdashline{#1}
  \noalign{\global\let\adl@draw\@dashdrawstore
           \vskip\belowrulesep}}
\title{StrucText-Eval: Evaluating Large Language Model's Reasoning Ability in Structure-Rich Text}

\pgfplotsset{compat=1.15}
\author{
Zhouhong Gu$^\heartsuit$\thanks{Equal Contribution}\ , 
Haoning Ye$^{\heartsuit*}$, 
Xingzhou Chen$^{\heartsuit}$, 
Zeyang Zhou$^\heartsuit$, \\
\bf
Hongwei Feng$^\heartsuit$\thanks{Corresponding authors.}\ , 
Yanghua Xiao$^{\heartsuit\clubsuit\dagger}$, \\
$^\heartsuit$Shanghai Key Laboratory of Data Science, School of Computer Science, Fudan University\\
$^\clubsuit$Fudan-Aishu Cognitive Intelligence Joint Research Center\\
\small \texttt{\{zhgu22, xzchen24\}@m.fudan.edu.cn} \\
\small \texttt{\{hnye19,zeyangzhou20,hwfeng,shawyh\}@fudan.edu.cn}
}

\makeatletter
\patchcmd{\hyper@makecurrent}{%
    \ifx\Hy@param\Hy@chapterstring
        \let\Hy@param\Hy@chapapp
    \fi
}{%
    \iftoggle{inappendix}{
        \@checkappendixparam{chapter}%
        \@checkappendixparam{section}%
        \@checkappendixparam{subsection}%
        \@checkappendixparam{subsubsection}%
        \@checkappendixparam{paragraph}%
        \@checkappendixparam{subparagraph}%
    }{}%
}{}{\errmessage{failed to patch}}
\newcommand*{\@checkappendixparam}[1]{%
    \def\@checkappendixparamtmp{#1}%
    \ifx\Hy@param\@checkappendixparamtmp
        \let\Hy@param\Hy@appendixstring
    \fi
}
\makeatletter
\newtoggle{inappendix}
\togglefalse{inappendix}
\apptocmd{\appendix}{\toggletrue{inappendix}}{}{\errmessage{failed to patch}}
\begin{document}
\maketitle
\begin{abstract}

The effective utilization of structured data, integral to corporate data strategies, has been challenged by the rise of large language models (LLMs) capable of processing unstructured information.
This shift prompts the question: can LLMs interpret structured data directly in its unstructured form?
We propose an automatic evaluation data generation method for assessing LLMs' reasoning capabilities on structure-rich text to explore this.
Our approach supports 8 structured languages and 29 tasks, generating data with adjustable complexity through controllable nesting and structural width.
We introduce StrucText-Eval, a benchmark containing 5,800 pre-generated and annotated samples designed to evaluate how well LLMs understand and reason through structured text.
StrucText-Eval is divided into two suites: a regular Test suite (3,712 samples) and a Test-Hard suite (2,088 samples), the latter emphasizing the gap between human and model performance on more complex tasks.
Experimental results show that while open-source LLMs achieve a maximum accuracy of 74.9\% on the standard dataset, their performance drops significantly to 45.8\% on the harder dataset.
In contrast, human participants reach an accuracy of 92.6\% on StrucText-Eval-Hard, highlighting LLMs' current limitations in handling intricate structural information.
The benchmark and generation codes are open sourced in \url{https://github.com/MikeGu721/StrucText-Eval}



\end{abstract}
\vspace{-5mm}
\section{Introduction}
\label{sec:intro}
Structured data, often represented by various structured languages such as JSON~\cite{pezoa2016foundations}, YAML~\cite{evans2001yaml}, ORG~\cite{orgModeManualHistory2023}, or Markdown~\cite{gruber2012markdown}, Latex~\cite{Lamport1985LatexA} etc., has consistently been central to corporate data strategies due to its ability to capture, store, and analyze essential information systematically.
The inherent benefits of structured data lie in its standardized format and high degree of organization, which facilitates efficient data querying and machine processing, clearly surpassing the inherent chaos of unstructured data.
However, with the advancement of large language models (LLMs)~\cite{achiam2023gpt,touvron2023llama,touvron2023llama2,sun2021ernie}, there has been a significant shift towards the effective utilization of unstructured data, attributed to the LLMs' capacity to comprehend and generate complex and nuanced semantics within such data~\cite{brown2020language}.
Considering that structured data can be directly presented in an unstructured format, it makes us wonder: \textbf{\textit{whether it is possible to rely on LLMs to interpret structured data through unstructured format directly}}.

Current LLM researchers have addressed their comprehension of structure-rich text of limited categories: 
Graphs~\cite{fatemi2023talk, perozzi2024let, guo2023gpt4graph, tang2023graphgpt, chen2023exploring}, 
Tables~\cite{sui2024table,campbell2003history,pasupat2015compositional}
and JSON~\cite{chen2024beyond, suzgun2022challenging}.
However, these categories do not encompass all potential use cases of structure-rich text.
For instance, scenarios requiring a direct understanding of articles in Latex or Markdown formats, data in YAML or ORG formats, or various custom-structured languages need to be adequately covered.
Moreover, existing benchmarks often rely on manually annotated data for evaluation, which limits the development of robust evaluation frameworks and potentially facilitates model cheating~\cite {zhou2023don}.

We propose a method for automatically generating evaluation data to assess models' capabilities in structure-rich text reasoning. 
This method is applied to 8 structured languages, as shown in Fig.~\ref{fig:taxonomy}, across 29 specific tasks, enabling data generation with controllable difficulty by adjusting the depth of structured nesting and the number of width and columns in the sample.
Based on this method, we further introduce the \textbf{\textit{Struc}}ture-Rich \textbf{\textit{Text}} \textbf{\textit{Eval}}uation Benchmark~(\textbf{StrucText-Eval}), a comprehensive benchmark with 5,800 pre-generated and annotated samples designed to evaluate the proficiency of LLMs in deciphering embedded structures within input text.
StrucText-Eval aims to evaluate whether LLMs understand raw structural tags, execute logical inferences based on the decoded semantics of these symbols, and organize their responses according to instruction requirements.
StrucText-Eval is divided into Test suite and Test-Hard suite, each consisting of 3,712 and 2,088 samples.
Considering that humans excel at understanding structured expressions, 
The Test-Hard suite features significantly longer questions, with an average length of 16,535 characters, and longest category contains 102,531 characters, which underscores the considerable gap between LLMs and human capabilities in comprehending and processing structured data.
The experimental results indicate that StrucText-Eval presents significant challenges in evaluating current LLMs' structured text processing capabilities.
While various open-sourced models achieve a maximum accuracy of 74.9\% under different prompting methods, their performance declines markedly to 45.8\% when tested on the more complex StrucText-Eval-Hard dataset. 
In contrast, human participants attain an accuracy of 92.6\% on StrucText-Eval-Hard, highlighting the limitations of existing LLMs in comprehending and reasoning through complex structural information.


\section{Related Work}
\label{sec:rtwk}
\begin{figure*}[t]

\centering
\fontsize{9pt}{8.6pt}\selectfont
  \begin{forest}
    for tree={
      align=center,
      edge+={thick, -{Stealth[]}},
      l sep'+=10pt,
      fork sep'=10pt,
    },
    forked edges,
    if level=0{
      inner xsep=0pt,
      tikz={\draw [thick] (.children first) -- (.children last);}
    }{},
    [Structure-Rich Texts
      [Structured Data 
      [Tabular [CSV]]
      [Tree [Custom Language]]]
      [Semi-Structured Data
        [Object Notation
          [JSON]
          [YAML]
          [XML]
        ]
        [Markup Language
          [Markdown]
          [LaTeX]
          [Org]
        ]
      ]
    ]
  \end{forest}
  
\caption{Taxonomy of Structure-Rich Texts covered in StrucText-Eval.}
\label{fig:taxonomy}
\vspace{-5mm}
\end{figure*}
\subsection{Structural Text Understanding Enhancements}

Recent efforts to enhance LLMs have focused on integrating external structures such as graphs, tool flows, and cross-domain representations to improve reasoning capabilities across various tasks.
For instance, ControlLLM utilizes tool graphs to decompose complex multimodal tasks, resulting in enhanced performance on image and audio processing tasks by leveraging the topological dependencies of tools \cite{1.pdf}.
Graph-based models like GraphGPT and BooG have shown promising results, with the former improving generalization across node classification and molecular tasks via graph instruction tuning \cite{7.pdf, 14.pdf}. At the same time, the latter employs virtual supernodes to unify graph structures across domains, fostering cross-domain task transferability \cite{2.pdf}.
Additionally, methods like RC2R demonstrate the effective combination of knowledge graphs and LLMs for domain-specific causal reasoning, particularly in financial risk propagation tasks \cite{4.pdf}.
These advancements highlight the benefits of embedding structural elements, from graph architectures to domain-specific knowledge graphs, within LLM frameworks to improve task-specific inference and reasoning.

\subsection{Structural Text Understanding Evaluation}

Evaluating LLMs' understanding of structured data has become increasingly critical, though benchmarks remain limited. GraphQA and Struc-Bench are key datasets that assess LLMs' reasoning over graph-structured data and tabular text, respectively, illustrating the models' varying capabilities based on input encoding \cite{fatemi2023talk, tang2023struc}.
More specialized benchmarks, such as TEMPTABQA, evaluate temporal reasoning in tabular data, while TableLLM tests LLMs' proficiency in handling complex document-based table manipulation tasks \cite{12.pdf, 20.pdf}. Other works, such as the evaluation of knowledge graph-based reasoning in complex time-series QA systems (JMFRN) \cite{5.pdf}, and privacy-oriented graph tasks in GHRatio \cite{3.pdf}, further explore how LLMs handle intricate, structure-rich information, shedding light on their performance across different structured data formats.

Our work diverges from prior research by focusing exclusively on structure-based inference, deliberately removing semantic content to challenge LLMs to reason purely from structural patterns. Unlike previous approaches that use structural data as supplementary input for classification or semantic tasks \cite{pasupat2015compositional, sui2024table}, we design semantically agnostic tasks requiring models to infer meaning solely from symbolic structures. Moreover, while earlier benchmarks emphasize graph reasoning or tabular information retrieval, our work extends to a broader spectrum of structure-rich text types, encompassing various input formats and more complex dependency-based inference tasks.

\section{StrucText-Eval Construction}
\label{sec:datasetconstr}

\subsection{Structure-Rich Texts Taxonomy}
To explore structure-rich texts comprehensively, we propose a dataset for eight structured data types, each categorized within a taxonomy depicted in Fig.~\ref{fig:taxonomy}.
This taxonomy encompasses both structured and semi-structured data formats.
The structured data types include Tree (\cite{cormen2022introduction}), Tabular (\cite{campbell2003history}), and Object Notation such as JSON (\cite{pezoa2016foundations}), YAML (\cite{evans2001yaml}), and XML (\cite{bray1998extensible}).
The semi-structured data types include Markup Languages like Markdown (\cite{gruber2012markdown}), LaTeX (\cite{Lamport1985LatexA}), and Org (\cite{orgModeManualHistory2023}).
Within StrucText-Eval, Tabular is stored in CSV format, whereas Tree is denoted by a custom format that nodes are represented as the string "xxx", connected with "->" and separated by "\textbackslash n". 
For examples encompassing all languages and tasks, please refer to Sec.~\ref{ap:samples} in the Appendix.


\begin{table}[t]
    \centering
    \resizebox{\columnwidth}{!}{
    \begin{tabular}{rrrcc}
    \toprule
     \#Sample & \#Reference & \#GroundTruth & Depth & Width    \\
     \midrule
     \multicolumn{5}{l}{\cellcolor{gray!20} \textit{StrucText-Eval-Test}} \\
     \midrule
     \textbf{3,712} & \textbf{804} & \textbf{47} & - & - \\
     1,856 & 582 & 19 & 1 & 1 \\
     1,856 & 1,026 & 74 & 2 & 1 \\
     \midrule
     \multicolumn{5}{l}{\cellcolor{gray!20} \textit{StrucText-Eval-Test-Hard}} \\
     \midrule
     
     \textbf{2,088} & \textbf{16,535} & \textbf{1,169} & - & - \\
    232 & 573 & 22 & 1 & 1 \\
    232 & 614 & 26 & 1 & 2 \\
    232 & 663 & 25 & 1 & 3 \\
    232 & 992 & 80 & 2 & 1 \\
    232 & 2,108 & 136 & 2 & 2 \\
    232 & 3,866 & 283 & 2 & 3 \\
    232 & 5,036 & 312 & 3 & 1 \\
    232 & 32,428 & 2,229 & 3 & 2 \\
    232 & 102,531 & 7,411 & 3 & 3 \\
     \bottomrule
    \end{tabular}
    }
    \caption{Statistics for StrucText-Eval test suite.}
    \label{tab:statistic}
\vspace{-5mm}
\end{table}

\begin{figure}[t]
\begin{center}
\includegraphics[width=\linewidth]{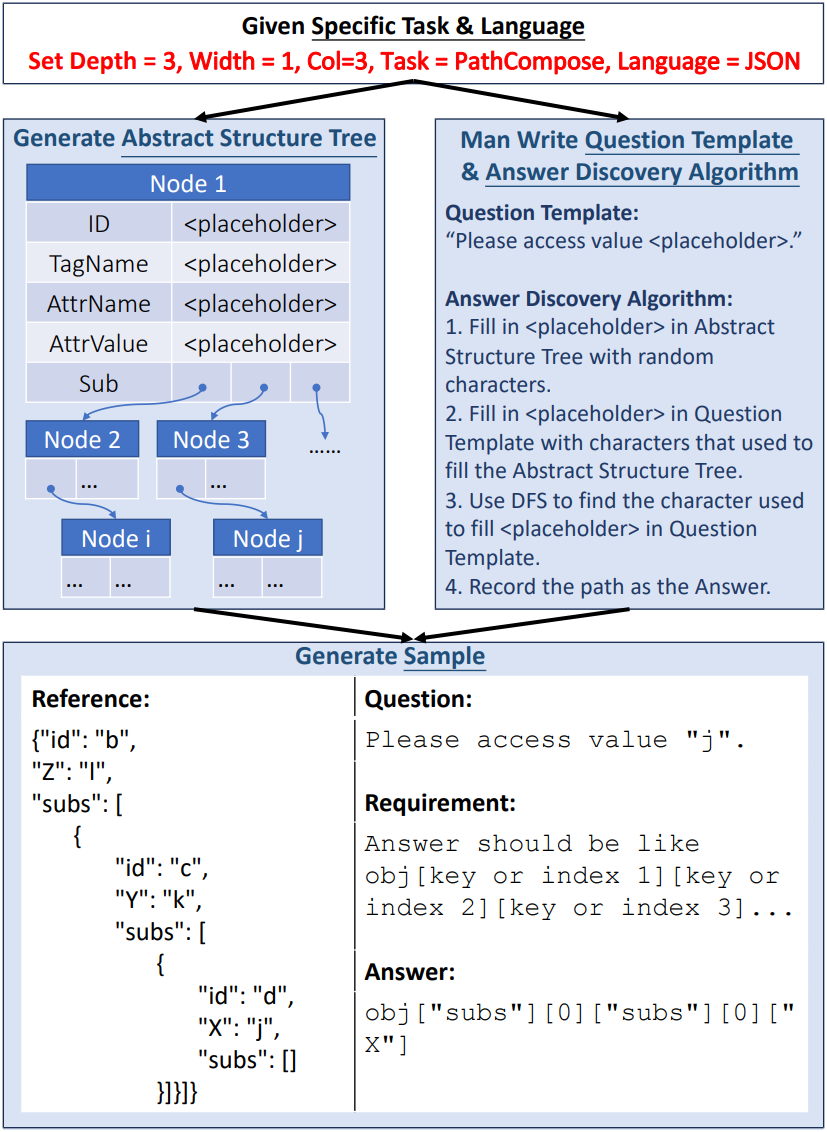}
\end{center}
\caption{
The illustration of the dataset generation process, the Json PathCompose task, is an example.
}
\label{fig:example}
\vspace{-5mm}
\end{figure}

\subsection{Dataset Generation}\label{sec:datagen}

An example of JSON's PathCompose is shown in Fig.~\ref{fig:example} to illustrate the dataset generation process.
The generation process mainly entails constructing an abstract structure tree, manually drafting question templates, and developing corresponding answer discovery algorithms.
The first step of the generation process is to define the complexity of the problem, characterized by depth, width, and column (Col), as well as its type, including task and language.
During the construction of the abstract tree, depth represents the depth of the tree, width indicates the number of children for each non-leaf node, and Col specifies the number of fields associated with each node. When constructing the question template, predefined templates are retrieved based on the specified task. Finally, during sample generation, the selected task is used to identify the corresponding ground truth according to specific rules, and both the abstract tree and the ground truth are translated into the selected language.

Eight task categories have been delineated for eight languages, as detailed in Fig.~\ref{tab:description}.
29 rules and question templates have been formulated for these tasks, with the specific rule templates detailed in Sec.~\ref{ap:rules} in the Appendix.
Each sample in the dataset comprises four main fields: ``Reference'', ``Question'', ``Requirement'' and ``Answer''.
We give examples for each language and task in Sec.~\ref{ap:samples} in the Appendix.

\begin{figure*}[t]
\centering
\begin{subfigure}{0.4\textwidth}

\centering
\resizebox{\columnwidth}{!}{
\begin{tikzpicture}[x=1.0cm,y=1.0cm,scale=1,baseline={(0,0)}]
\def\innername{\tiny{StrucText-Eval}}
\def\innerradius{1.7cm}
\def\outerradius{2.6cm}
\pgfmathsetlengthmacro{\centerradius}{(\outerradius + \innerradius)/2}
\pgfmathsetlengthmacro{\donutcenter}{\innerradius/2}
\donutchart{
  8/pppurple/PC, 8/rrred/ND, 8/brown!50/JO,
  8/olive!50/TH, 16/ooorange/ST, 8/cccyan/TR,
  8/pppurple/PC, 8/bbblue/PW, 8/lllightorange/SY, 16/cccyan/TR,
  8/pppurple/PC, 8/bbblue/PW, 8/lllightorange/SY, 16/cccyan/TR,
  8/lllightorange/SY, 16/cccyan/TR,
  8/bbblue/PW, 16/cccyan/TR,
  8/bbblue/PW, 16/cccyan/TR,
  8/bbblue/PW, 16/cccyan/TR
}{2}

\def\innerradius{0.7cm}
\def\outerradius{1.7cm}
\pgfmathsetlengthmacro{\centerradius}{(\outerradius + \innerradius)/2}
\pgfmathsetlengthmacro{\donutcenter}{\innerradius/2}
\donutchart{
  24/bblue/Tree,
  32/olive!75/Tabular,
  40/rred/JSON,
  40/oorange/YAML,
  24/ggreen/XML,
  24/ppurple/Markdown,
  24/ccyan/ORG,
  24/llightorange/LaTeX
}{1}
\end{tikzpicture}
}
\caption{Benchmark Decomposition }
\label{fig:datamisc}

\end{subfigure}\hfill
\begin{subfigure}{0.58\textwidth}

\resizebox{\textwidth}{!}{
\begin{tabular}{>{\centering\arraybackslash}m{2.5cm}cm{10cm}}
\toprule
\textbf{Task Name} & \textbf{Abbr.} & \textbf{Task Description} \\ \hline
Syntax & SY& Focuses on detecting structural errors in data formats such as JSON, XML, and YAML. \\ \hline
PathWalk &PW& Focuses on extracting specific sections or subsections from structured documents such as org, LaTeX, or markdown files. \\ \hline
TextRetrieval &TR& Assesses the ability to extract specific information from various document formats, including text content and image filenames. \\ \hline
Statistic &ST& Concentrates on statistical queries to calculate the number of employees meeting specific salary conditions. \\ \hline
Join &JO& Assesses the ability to filter data sets that meet specific criteria by combining multiple tables in a database through SQL queries. \\ \hline
Tree.Height &TH& Evaluates calculating the height of the longest path from the root node to any leaf node in a tree structure. \\ \hline
Node.Depth &ND& Assesses the depth of any node in a tree structure relative to the root node. \\ \hline
PathCompose &PC& Evaluates reasoning of paths and multi-level data indexing within hierarchical or tree-like structures. \\ \bottomrule
\end{tabular}
}
\caption{Descriptions of tasks for evaluating structured data understanding in large language models}
\label{tab:description}
\end{subfigure}

\caption{The tasks within StrucText-Eval and their description.}
\label{fig:task_desc}

\vspace{-5mm}
\end{figure*}

\subsection{Statistic Information}
StrucText-Eval has assembled two datasets.
StrucText-Eval-Test comprises 3,712 samples, and StrucText-Eval-Test-Hard comprises 2,088 samples, each of the 29 specific tasks for eight languages as depicted in Fig.~\ref{fig:datamisc}.
Detailed statistics regarding the number of samples, lengths, and complexity levels across all tasks, languages, and difficulties are detailed in Tab.~\ref{tab:statistic}.

\section{Experiment Setup}
\label{sec:setup}
To evaluate LLMs' current capability of processing structure-rich text and executing dependent inference, we conducted a series of experiments using StrucText-Eval in various settings.
Our study utilizes both prompt-based and finetuning methods to analyze the performance variations.

\subsection{Models}

We tested six Open-Source LLMs in both StrucText-Eval Test and Test-Hard Suite, and we use the short name (in the bracket) of these LLMs in the experiments:
\textbf{Qwen/Qwen2-7B-Instruct} (Qwen-2-7B),
\textbf{Qwen/Qwen2-72B-Instruct} (Qwen-2-72B),
\textbf{meta-llama/Meta-Llama-3.1-8B-Instruct} (Llama-3.1-8B),
\textbf{meta-llama/Meta-Llama-3.1-72B-Instruct} (Llama-3.1-70B),
\textbf{meta-llama/Meta-Llama-3.1-405B-Instruct}  (Llama-3.1-405B),
\textbf{mistralai/Mistral-7B-Instruct-v0.2} (Mistral-0.2-7B)

Considering the huge expense of using an API-based model, we only tested six Close-Source LLMs in StrucText-Eval-Hard:
\textbf{gpt-4o-2024-08-06} (gpt-4o),
\textbf{gpt-4o-mini-2024-07-18} (gpt-4o-mini),
\textbf{gemini-1.5-pro}(gemini-1.5-pro),
\textbf{gemini-1.5-flash}(gemini-1.5-flash),
\textbf{GLM-4-Plus} (glm-4-plus),
\textbf{GLM-4-Flash} (glm-4-flash).

\subsection{Prompt-based Method}
We also evaluated the impact of different prompt designs on the performance of LLMs by utilizing six distinct prompt configurations in the main experiments.
Detailed implementation of these prompts can be found in Sec.~\ref{ap:prompt} in the Appendix.
The six primary prompt settings are as follows:

\texttt{\textbf{Naive}}: This configuration involves a straightforward input of ``Context'', ``Question'', and ``Options'' into the LLMs to generate responses.
\texttt{\textbf{Self-Chain-of-Thought (Self-CoT)}}~\cite{kojima2022large}: This approach incorporates a step-by-step reasoning prompt to guide the model through logical reasoning.
\texttt{\textbf{Plan-and-Solve CoT (PS-CoT)}}~\cite{wang2023plan}: This method emphasizes problem decomposition before solving, encouraging the model to first break down the problem before generating a solution.
\texttt{\textbf{With Hint (w/ hint)}}: In this setting, manually curated hints are provided to the model to observe its performance when additional information is injected. 
Since this approach introduces supplementary data, it is delineated by a dashed line from other methods in Table~\ref{tab:api_overall}.
\texttt{\textbf{Few-Shot Demonstration}}: involves appending few training data directly to the prompt. 
The \texttt{\textbf{Simple Few-Shot Demonstration}} uses only the shortest examples from the training set as few-shot demonstrations.

\subsection{Evaluation Method}
We use the RougeL metric~\cite{lin2004rouge} to assess the degree of character-level similarity between model outputs in the main content of this paper.
Sometimes, the task requires the LLM to generate the entire reasoning path leading to the answer, which results in high RougeL scores.
So, we assign a score of 0 if the RougeL score falls below 0.75.

\begin{figure}
    \centering
    \includegraphics[width=\linewidth]{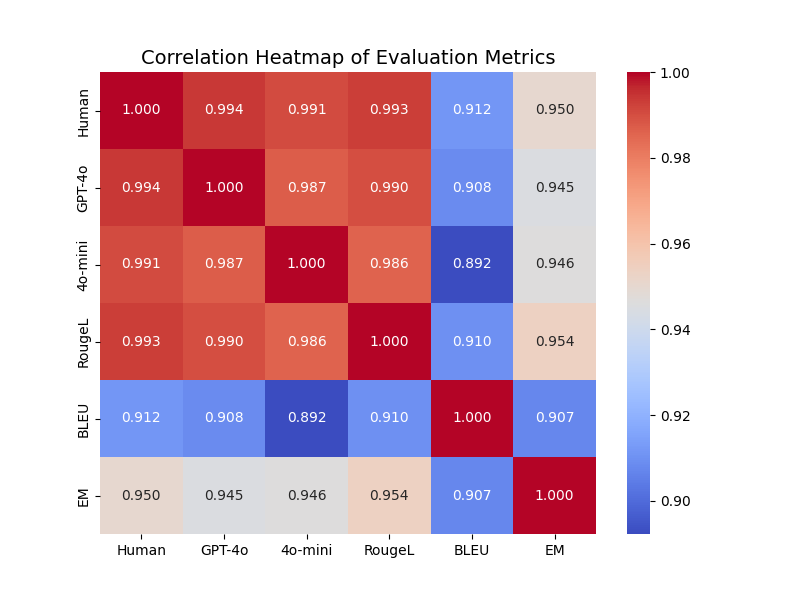}
    \caption{Correlation between different evaluation metrics.}
    \label{fig:cor_metrics}
\vspace{-5mm}
\end{figure}

Additionally, we present the results of other evaluation metrics, including LLM-as-Judge-Score~\cite{zheng2023judging}, BLEU~\cite{papineni2002bleu}, and Exact Match, in Tab.~\ref{tab:other_metrics} in the Appendix.
Furthermore, we conduct a consistency analysis across these metrics compared to human judgments, as shown in Fig.~\ref{fig:cor_metrics}.

\section{Analysis}
\label{sec:analysis}

\definecolor{lightblue}{RGB}{173, 216, 230}
\definecolor{lightpink}{RGB}{255, 182, 193}  
\definecolor{lightgray}{RGB}{211, 211, 211}  
\definecolor{lightgreen}{RGB}{144, 238, 144}
\definecolor{lightyellow}{RGB}{255, 255, 200}  
\definecolor{lightred}{RGB}{255, 182, 193}
\definecolor{lightpurple}{RGB}{216, 191, 216}

\definecolor{darkblue}{RGB}{123, 166, 180}  
\definecolor{darkpink}{RGB}{199, 21, 133}   
\definecolor{darkgray}{RGB}{169, 169, 169}  
\definecolor{darkgreen}{RGB}{94, 188, 94}    
\definecolor{darkyellow}{RGB}{218, 165, 32}   
\definecolor{darkred}{RGB}{205, 92, 92}     
\definecolor{darkpurple}{RGB}{166, 141, 166}

\begin{table*}[t]
\centering
{\fontsize{6.8pt}{8.6pt}\selectfont
\setlength{\tabcolsep}{2.6pt} 
\begin{tabular}{llccccccccccccccccc}
\toprule
\multirow{2}{*}{\textbf{Model}}&
\multirow{2}{*}{\textbf{Prompt}} &
\multicolumn{8}{c}{\textbf{Languages}} &
\multicolumn{8}{c}{\textbf{Tasks}} &
\multirow{2}{*}{\textbf{all}}\\
\cline{3-10}
 \cdashlinelr{11-18} 
 && \textbf{JSON} & \textbf{LaTeX} & \textbf{Md.} & \textbf{ORG} & \textbf{CSV} & \textbf{Tree} & \textbf{XML} & \textbf{YAML}
 & \textbf{PC} & \textbf{PW} & \textbf{SY} & \textbf{TR} & \textbf{JO} & \textbf{ST} & \textbf{ND} & \textbf{TH} &  \\ 
\midrule
\multirow{4}{*}{Qwen2-7B} & \cellcolor{darkblue!25}Base & \cellcolor{lightblue!25}70.4 & \cellcolor{lightblue!25}68.8 & \cellcolor{lightblue!25}68.0 & \cellcolor{lightblue!25}54.5 & \cellcolor{lightblue!25}83.5 & \cellcolor{lightblue!25}68.9 & \cellcolor{lightblue!25}57.6 & \cellcolor{lightblue!25}68.5 & \cellcolor{lightblue!25}48.5 & \cellcolor{lightblue!25}74.2 & \cellcolor{lightblue!25}49.2 & \cellcolor{lightblue!25}72.4 & \cellcolor{lightblue!25}79.5 & \cellcolor{lightblue!25}78.4 & \cellcolor{lightblue!25}47.7 & \cellcolor{lightblue!25}93.2 & \cellcolor{darkblue!25}30.0\\
 & \cellcolor{darkblue!25}Self-CoT & \cellcolor{lightblue!25}12.8 & \cellcolor{lightblue!25}1.5 & \cellcolor{lightblue!25}1.5 & \cellcolor{lightblue!25}9.1 & \cellcolor{lightblue!25}29.0 & \cellcolor{lightblue!25}4.5 & \cellcolor{lightblue!25}3.6 & \cellcolor{lightblue!25}3.5 & \cellcolor{lightblue!25}4.5 & \cellcolor{lightblue!25}6.4 & \cellcolor{lightblue!25}6.1 & \cellcolor{lightblue!25}8.1 & \cellcolor{lightblue!25}27.3 & \cellcolor{lightblue!25}26.1 & \cellcolor{lightblue!25}2.3 & \cellcolor{lightblue!25}6.8 & \cellcolor{darkblue!25}17.2\\
 & \cellcolor{darkblue!25}PS-CoT & \cellcolor{lightblue!25}31.7 & \cellcolor{lightblue!25}31.7 & \cellcolor{lightblue!25}19.4 & \cellcolor{lightblue!25}20.1 & \cellcolor{lightblue!25}67.0 & \cellcolor{lightblue!25}36.4 & \cellcolor{lightblue!25}25.8 & \cellcolor{lightblue!25}24.9 & \cellcolor{lightblue!25}9.8 & \cellcolor{lightblue!25}19.8 & \cellcolor{lightblue!25}32.6 & \cellcolor{lightblue!25}34.1 & \cellcolor{lightblue!25}63.6 & \cellcolor{lightblue!25}60.2 & \cellcolor{lightblue!25}25.0 & \cellcolor{lightblue!25}72.7 & \cellcolor{darkblue!25}29.1\\
\cdashlinelr{2-19}
 & \cellcolor{darkblue!25}w/ Hint & \cellcolor{lightblue!25}70.8 & \cellcolor{lightblue!25}66.1 & \cellcolor{lightblue!25}66.5 & \cellcolor{lightblue!25}58.1 & \cellcolor{lightblue!25}85.2 & \cellcolor{lightblue!25}56.8 & \cellcolor{lightblue!25}55.2 & \cellcolor{lightblue!25}70.2 & \cellcolor{lightblue!25}43.9 & \cellcolor{lightblue!25}72.3 & \cellcolor{lightblue!25}43.2 & \cellcolor{lightblue!25}75.3 & \cellcolor{lightblue!25}86.4 & \cellcolor{lightblue!25}77.3 & \cellcolor{lightblue!25}45.5 & \cellcolor{lightblue!25}65.9 & \cellcolor{darkblue!25}44.0\\
\cdashlinelr{1-19}
\multirow{4}{*}{Qwen2-72B} & \cellcolor{darkpurple!25}Base & \cellcolor{lightpurple!25}85.8 & \cellcolor{lightpurple!25}\underline{73.7} & \cellcolor{lightpurple!25}75.1 & \cellcolor{lightpurple!25}67.1 & \cellcolor{lightpurple!25}92.6 & \cellcolor{lightpurple!25}86.4 & \cellcolor{lightpurple!25}71.2 & \cellcolor{lightpurple!25}82.7 & \cellcolor{lightpurple!25}80.3 & \cellcolor{lightpurple!25}81.5 & \cellcolor{lightpurple!25}62.9 & \cellcolor{lightpurple!25}80.8 & \cellcolor{lightpurple!25}90.9 & \cellcolor{lightpurple!25}90.9 & \cellcolor{lightpurple!25}77.3 & \cellcolor{lightpurple!25}\underline{95.5} & \cellcolor{darkpurple!25}42.6\\
 & \cellcolor{darkpurple!25}Self-CoT & \cellcolor{lightpurple!25}85.4 & \cellcolor{lightpurple!25}69.9 & \cellcolor{lightpurple!25}70.8 & \cellcolor{lightpurple!25}65.2 & \cellcolor{lightpurple!25}95.5 & \cellcolor{lightpurple!25}\textbf{90.2} & \cellcolor{lightpurple!25}79.5 & \cellcolor{lightpurple!25}89.7 & \cellcolor{lightpurple!25}78.8 & \cellcolor{lightpurple!25}77.1 & \cellcolor{lightpurple!25}81.1 & \cellcolor{lightpurple!25}81.7 & \cellcolor{lightpurple!25}90.9 & \cellcolor{lightpurple!25}\underline{95.5} & \cellcolor{lightpurple!25}84.1 & \cellcolor{lightpurple!25}\underline{95.5} & \cellcolor{darkpurple!25}51.0\\
 & \cellcolor{darkpurple!25}PS-CoT & \cellcolor{lightpurple!25}89.5 & \cellcolor{lightpurple!25}70.1 & \cellcolor{lightpurple!25}68.9 & \cellcolor{lightpurple!25}61.7 & \cellcolor{lightpurple!25}92.0 & \cellcolor{lightpurple!25}84.8 & \cellcolor{lightpurple!25}81.1 & \cellcolor{lightpurple!25}93.4 & \cellcolor{lightpurple!25}76.5 & \cellcolor{lightpurple!25}77.6 & \cellcolor{lightpurple!25}87.9 & \cellcolor{lightpurple!25}80.8 & \cellcolor{lightpurple!25}81.8 & \cellcolor{lightpurple!25}93.2 & \cellcolor{lightpurple!25}\underline{86.4} & \cellcolor{lightpurple!25}\textbf{97.7} & \cellcolor{darkpurple!25}65.3\\
\cdashlinelr{2-19}
 & \cellcolor{darkpurple!25}w/ Hint & \cellcolor{lightpurple!25}90.0 & \cellcolor{lightpurple!25}72.5 & \cellcolor{lightpurple!25}\textbf{79.1} & \cellcolor{lightpurple!25}\underline{68.6} & \cellcolor{lightpurple!25}94.9 & \cellcolor{lightpurple!25}81.1 & \cellcolor{lightpurple!25}72.7 & \cellcolor{lightpurple!25}90.8 & \cellcolor{lightpurple!25}81.1 & \cellcolor{lightpurple!25}\underline{84.0} & \cellcolor{lightpurple!25}77.3 & \cellcolor{lightpurple!25}82.4 & \cellcolor{lightpurple!25}95.5 & \cellcolor{lightpurple!25}92.0 & \cellcolor{lightpurple!25}72.7 & \cellcolor{lightpurple!25}86.4 & \cellcolor{darkpurple!25}49.4\\
\cdashlinelr{1-19}
\multirow{4}{*}{Llama-3.1-8B} & \cellcolor{darkgray!25}Base & \cellcolor{lightgray!25}43.9 & \cellcolor{lightgray!25}64.6 & \cellcolor{lightgray!25}49.3 & \cellcolor{lightgray!25}48.3 & \cellcolor{lightgray!25}42.6 & \cellcolor{lightgray!25}50.0 & \cellcolor{lightgray!25}26.5 & \cellcolor{lightgray!25}46.9 & \cellcolor{lightgray!25}30.3 & \cellcolor{lightgray!25}49.4 & \cellcolor{lightgray!25}1.5 & \cellcolor{lightgray!25}61.0 & \cellcolor{lightgray!25}11.4 & \cellcolor{lightgray!25}45.5 & \cellcolor{lightgray!25}22.7 & \cellcolor{lightgray!25}79.5 & \cellcolor{darkgray!25}21.3\\
 & \cellcolor{darkgray!25}Self-CoT & \cellcolor{lightgray!25}52.2 & \cellcolor{lightgray!25}40.6 & \cellcolor{lightgray!25}49.2 & \cellcolor{lightgray!25}39.0 & \cellcolor{lightgray!25}66.5 & \cellcolor{lightgray!25}43.2 & \cellcolor{lightgray!25}36.6 & \cellcolor{lightgray!25}55.2 & \cellcolor{lightgray!25}40.9 & \cellcolor{lightgray!25}40.2 & \cellcolor{lightgray!25}39.7 & \cellcolor{lightgray!25}53.0 & \cellcolor{lightgray!25}77.3 & \cellcolor{lightgray!25}65.9 & \cellcolor{lightgray!25}52.3 & \cellcolor{lightgray!25}36.4 & \cellcolor{darkgray!25}48.5\\
 & \cellcolor{darkgray!25}PS-CoT & \cellcolor{lightgray!25}45.8 & \cellcolor{lightgray!25}18.7 & \cellcolor{lightgray!25}34.0 & \cellcolor{lightgray!25}32.8 & \cellcolor{lightgray!25}64.0 & \cellcolor{lightgray!25}63.1 & \cellcolor{lightgray!25}44.6 & \cellcolor{lightgray!25}41.3 & \cellcolor{lightgray!25}48.8 & \cellcolor{lightgray!25}50.5 & \cellcolor{lightgray!25}44.7 & \cellcolor{lightgray!25}32.8 & \cellcolor{lightgray!25}69.8 & \cellcolor{lightgray!25}56.8 & \cellcolor{lightgray!25}64.3 & \cellcolor{lightgray!25}62.8 & \cellcolor{darkgray!25}55.9\\
\cdashlinelr{2-19}
 & \cellcolor{darkgray!25}w/ Hint & \cellcolor{lightgray!25}44.9 & \cellcolor{lightgray!25}62.2 & \cellcolor{lightgray!25}55.9 & \cellcolor{lightgray!25}48.1 & \cellcolor{lightgray!25}29.0 & \cellcolor{lightgray!25}54.5 & \cellcolor{lightgray!25}30.5 & \cellcolor{lightgray!25}51.4 & \cellcolor{lightgray!25}31.8 & \cellcolor{lightgray!25}45.4 & \cellcolor{lightgray!25}9.1 & \cellcolor{lightgray!25}63.4 & \cellcolor{lightgray!25}2.3 & \cellcolor{lightgray!25}22.7 & \cellcolor{lightgray!25}38.6 & \cellcolor{lightgray!25}90.9 & \cellcolor{darkgray!25}26.9\\
\cdashlinelr{1-19}
\multirow{4}{*}{Llama-3.1-70B} & \cellcolor{darkgreen!25}Base & \cellcolor{lightgreen!25}\underline{93.8} & \cellcolor{lightgreen!25}70.9 & \cellcolor{lightgreen!25}69.8 & \cellcolor{lightgreen!25}62.8 & \cellcolor{lightgreen!25}72.7 & \cellcolor{lightgreen!25}51.5 & \cellcolor{lightgreen!25}78.7 & \cellcolor{lightgreen!25}88.8 & \cellcolor{lightgreen!25}81.8 & \cellcolor{lightgreen!25}75.4 & \cellcolor{lightgreen!25}82.6 & \cellcolor{lightgreen!25}81.0 & \cellcolor{lightgreen!25}72.7 & \cellcolor{lightgreen!25}59.1 & \cellcolor{lightgreen!25}47.7 & \cellcolor{lightgreen!25}43.2 & \cellcolor{darkgreen!25}50.8\\
 & \cellcolor{darkgreen!25}Self-CoT & \cellcolor{lightgreen!25}93.6 & \cellcolor{lightgreen!25}71.4 & \cellcolor{lightgreen!25}69.7 & \cellcolor{lightgreen!25}54.8 & \cellcolor{lightgreen!25}96.0 & \cellcolor{lightgreen!25}84.1 & \cellcolor{lightgreen!25}87.1 & \cellcolor{lightgreen!25}\underline{95.9} & \cellcolor{lightgreen!25}\underline{88.6} & \cellcolor{lightgreen!25}67.9 & \cellcolor{lightgreen!25}86.4 & \cellcolor{lightgreen!25}\textbf{85.7} & \cellcolor{lightgreen!25}\underline{97.7} & \cellcolor{lightgreen!25}93.2 & \cellcolor{lightgreen!25}77.3 & \cellcolor{lightgreen!25}\textbf{97.7} & \cellcolor{darkgreen!25}\textbf{76.7}\\
 & \cellcolor{darkgreen!25}PS-CoT & \cellcolor{lightgreen!25}\textbf{94.5} & \cellcolor{lightgreen!25}68.7 & \cellcolor{lightgreen!25}72.7 & \cellcolor{lightgreen!25}61.7 & \cellcolor{lightgreen!25}93.7 & \cellcolor{lightgreen!25}83.2 & \cellcolor{lightgreen!25}\underline{93.9} & \cellcolor{lightgreen!25}\textbf{98.5} & \cellcolor{lightgreen!25}\textbf{90.8} & \cellcolor{lightgreen!25}77.0 & \cellcolor{lightgreen!25}\textbf{93.9} & \cellcolor{lightgreen!25}84.2 & \cellcolor{lightgreen!25}93.2 & \cellcolor{lightgreen!25}90.9 & \cellcolor{lightgreen!25}81.8 & \cellcolor{lightgreen!25}90.9 & \cellcolor{darkgreen!25}72.9\\
\cdashlinelr{2-19}
 & \cellcolor{darkgreen!25}w/ Hint & \cellcolor{lightgreen!25}93.6 & \cellcolor{lightgreen!25}\textbf{73.9} & \cellcolor{lightgreen!25}\underline{77.4} & \cellcolor{lightgreen!25}\textbf{71.6} & \cellcolor{lightgreen!25}72.7 & \cellcolor{lightgreen!25}74.2 & \cellcolor{lightgreen!25}80.4 & \cellcolor{lightgreen!25}93.6 & \cellcolor{lightgreen!25}\underline{88.6} & \cellcolor{lightgreen!25}\textbf{84.9} & \cellcolor{lightgreen!25}84.1 & \cellcolor{lightgreen!25}83.5 & \cellcolor{lightgreen!25}70.5 & \cellcolor{lightgreen!25}60.2 & \cellcolor{lightgreen!25}65.9 & \cellcolor{lightgreen!25}75.0 & \cellcolor{darkgreen!25}58.4\\
\cdashlinelr{1-19}
\multirow{4}{*}{Llama-3.1-405B} & \cellcolor{darkyellow!25}Base & \cellcolor{lightyellow!25}82.0 & \cellcolor{lightyellow!25}62.9 & \cellcolor{lightyellow!25}70.0 & \cellcolor{lightyellow!25}60.9 & \cellcolor{lightyellow!25}\underline{96.6} & \cellcolor{lightyellow!25}65.9 & \cellcolor{lightyellow!25}61.5 & \cellcolor{lightyellow!25}78.1 & \cellcolor{lightyellow!25}74.2 & \cellcolor{lightyellow!25}69.4 & \cellcolor{lightyellow!25}32.6 & \cellcolor{lightyellow!25}82.4 & \cellcolor{lightyellow!25}\underline{97.7} & \cellcolor{lightyellow!25}94.3 & \cellcolor{lightyellow!25}45.5 & \cellcolor{lightyellow!25}79.5 & \cellcolor{darkyellow!25}38.3\\
 & \cellcolor{darkyellow!25}Self-CoT & \cellcolor{lightyellow!25}87.7 & \cellcolor{lightyellow!25}62.2 & \cellcolor{lightyellow!25}74.2 & \cellcolor{lightyellow!25}62.2 & \cellcolor{lightyellow!25}95.5 & \cellcolor{lightyellow!25}75.8 & \cellcolor{lightyellow!25}78.5 & \cellcolor{lightyellow!25}90.8 & \cellcolor{lightyellow!25}87.9 & \cellcolor{lightyellow!25}73.2 & \cellcolor{lightyellow!25}63.6 & \cellcolor{lightyellow!25}83.4 & \cellcolor{lightyellow!25}\textbf{100.0} & \cellcolor{lightyellow!25}90.9 & \cellcolor{lightyellow!25}59.1 & \cellcolor{lightyellow!25}88.6 & \cellcolor{darkyellow!25}67.1\\
 & \cellcolor{darkyellow!25}PS-CoT & \cellcolor{lightyellow!25}84.5 & \cellcolor{lightyellow!25}67.4 & \cellcolor{lightyellow!25}76.0 & \cellcolor{lightyellow!25}66.7 & \cellcolor{lightyellow!25}92.0 & \cellcolor{lightyellow!25}\underline{86.7} & \cellcolor{lightyellow!25}\textbf{94.7} & \cellcolor{lightyellow!25}94.7 & \cellcolor{lightyellow!25}88.3 & \cellcolor{lightyellow!25}79.1 & \cellcolor{lightyellow!25}\underline{93.2} & \cellcolor{lightyellow!25}81.1 & \cellcolor{lightyellow!25}\underline{97.7} & \cellcolor{lightyellow!25}85.2 & \cellcolor{lightyellow!25}\textbf{90.9} & \cellcolor{lightyellow!25}88.6 & \cellcolor{darkyellow!25}\underline{74.9}\\
\cdashlinelr{2-19}
 & \cellcolor{darkyellow!25}w/ Hint & \cellcolor{lightyellow!25}85.4 & \cellcolor{lightyellow!25}68.3 & \cellcolor{lightyellow!25}75.1 & \cellcolor{lightyellow!25}66.7 & \cellcolor{lightyellow!25}\textbf{98.3} & \cellcolor{lightyellow!25}70.5 & \cellcolor{lightyellow!25}74.5 & \cellcolor{lightyellow!25}87.2 & \cellcolor{lightyellow!25}74.2 & \cellcolor{lightyellow!25}78.0 & \cellcolor{lightyellow!25}59.1 & \cellcolor{lightyellow!25}\underline{84.9} & \cellcolor{lightyellow!25}\underline{97.7} & \cellcolor{lightyellow!25}\textbf{97.7} & \cellcolor{lightyellow!25}50.0 & \cellcolor{lightyellow!25}84.1 & \cellcolor{darkyellow!25}46.5\\
\cdashlinelr{1-19}
\multirow{4}{*}{Mistral-7B} & \cellcolor{darkred!25}Base & \cellcolor{lightred!25}32.5 & \cellcolor{lightred!25}42.1 & \cellcolor{lightred!25}44.9 & \cellcolor{lightred!25}40.2 & \cellcolor{lightred!25}9.1 & \cellcolor{lightred!25}4.5 & \cellcolor{lightred!25}14.8 & \cellcolor{lightred!25}33.5 & \cellcolor{lightred!25}6.1 & \cellcolor{lightred!25}30.8 & \cellcolor{lightred!25}0.0 & \cellcolor{lightred!25}47.7 & \cellcolor{lightred!25}0.0 & \cellcolor{lightred!25}6.8 & \cellcolor{lightred!25}0.0 & \cellcolor{lightred!25}0.0 & \cellcolor{darkred!25}11.3\\
 & \cellcolor{darkred!25}Self-CoT & \cellcolor{lightred!25}56.5 & \cellcolor{lightred!25}35.1 & \cellcolor{lightred!25}40.3 & \cellcolor{lightred!25}36.6 & \cellcolor{lightred!25}34.7 & \cellcolor{lightred!25}15.9 & \cellcolor{lightred!25}33.7 & \cellcolor{lightred!25}54.3 & \cellcolor{lightred!25}28.8 & \cellcolor{lightred!25}49.2 & \cellcolor{lightred!25}64.4 & \cellcolor{lightred!25}43.9 & \cellcolor{lightred!25}6.8 & \cellcolor{lightred!25}23.9 & \cellcolor{lightred!25}13.6 & \cellcolor{lightred!25}13.6 & \cellcolor{darkred!25}8.1\\
 & \cellcolor{darkred!25}PS-CoT & \cellcolor{lightred!25}43.9 & \cellcolor{lightred!25}19.7 & \cellcolor{lightred!25}22.9 & \cellcolor{lightred!25}15.6 & \cellcolor{lightred!25}14.8 & \cellcolor{lightred!25}18.2 & \cellcolor{lightred!25}34.6 & \cellcolor{lightred!25}44.1 & \cellcolor{lightred!25}18.9 & \cellcolor{lightred!25}30.6 & \cellcolor{lightred!25}56.8 & \cellcolor{lightred!25}29.1 & \cellcolor{lightred!25}22.7 & \cellcolor{lightred!25}6.8 & \cellcolor{lightred!25}13.6 & \cellcolor{lightred!25}22.7 & \cellcolor{darkred!25}19.5\\
\cdashlinelr{2-19}
 & \cellcolor{darkred!25}w/ Hint & \cellcolor{lightred!25}34.6 & \cellcolor{lightred!25}39.4 & \cellcolor{lightred!25}52.7 & \cellcolor{lightred!25}40.5 & \cellcolor{lightred!25}10.2 & \cellcolor{lightred!25}6.8 & \cellcolor{lightred!25}12.7 & \cellcolor{lightred!25}36.5 & \cellcolor{lightred!25}9.8 & \cellcolor{lightred!25}34.3 & \cellcolor{lightred!25}0.0 & \cellcolor{lightred!25}48.9 & \cellcolor{lightred!25}0.0 & \cellcolor{lightred!25}8.0 & \cellcolor{lightred!25}0.0 & \cellcolor{lightred!25}0.0 & \cellcolor{darkred!25}10.6\\

\bottomrule
\end{tabular}}
\caption{RougeL score for open sourced LLMs' performance. 
\textbf{Bolded} text represent the best performance in the column.
\underline{Underlined} text represent the second best performance in the column.
}
\label{tab:api_overall}
\vspace{-5mm}
\end{table*}

\begin{table}[t]
\resizebox{\columnwidth}{!}{
\begin{tabular}{lcccc}
\toprule
\multirow{2}{*}{\textbf{Model}} & \multicolumn{4}{c}{\textbf{Prompt}} \\
      & Base & w/ Hint & 3-Shot & Simple 3-Shot \\
     \hline
     \multirow{1}{*}{GPT-4o-Turbo}    & \cellcolor{lightblue!25}\textbf{51.1} & \cellcolor{lightblue!25}\textbf{54.2} & \cellcolor{lightblue!25}\textbf{69.5} &\cellcolor{lightblue!25}\underline{49.7}\\
     \multirow{1}{*}{GPT-4o-Mini}    & \cellcolor{lightpurple!25}39.3 & \cellcolor{lightpurple!25}47.7 & \cellcolor{lightpurple!25}\underline{65.6} &\cellcolor{lightpurple!25}39.9\\
     \multirow{1}{*}{Gemini1.5-Pro}    & \cellcolor{lightgray!25}11.2 & \cellcolor{lightgray!25}15.7 & \cellcolor{lightgray!25}53.0 &\cellcolor{lightgray!25}12.5\\
     \multirow{1}{*}{Gemini1.5-Pro-Flash}    & \cellcolor{lightgreen!25}12.9 & \cellcolor{lightgreen!25}12.9 & \cellcolor{lightgreen!25}38.3 &\cellcolor{lightgreen!25}11.9\\
     \multirow{1}{*}{GLM-4-Plus}    & \cellcolor{lightyellow!25}\underline{47.3} & \cellcolor{lightyellow!25}\underline{50.9} & \cellcolor{lightyellow!25}65.8 &\cellcolor{lightyellow!25}\textbf{51.7}\\
     \multirow{1}{*}{GLM-4-Flash}    & \cellcolor{lightred!25}40.9 & \cellcolor{lightred!25}47.8 & \cellcolor{lightred!25}55.2 &\cellcolor{lightred!25}41.7\\
     
     \hline
     \hline
     \multirow{1}{*}{QWen-2-7B}    & \cellcolor{lightblue!25}29.6 & \cellcolor{lightblue!25}35.0 & \cellcolor{lightblue!25}51.9 &\cellcolor{lightblue!25}30.0\\
     \multirow{1}{*}{QWen-2-72B}    & \cellcolor{lightpurple!25}\underline{42.5} & \cellcolor{lightpurple!25}\underline{45.3} & \cellcolor{lightpurple!25}\textbf{61.4} &\cellcolor{lightpurple!25}36.2\\
     \multirow{1}{*}{Llama-3.1-8B}    & \cellcolor{lightgray!25}22.3 & \cellcolor{lightgray!25}26.7 & \cellcolor{lightgray!25}33.7 &\cellcolor{lightgray!25}34.2\\
     \multirow{1}{*}{Llama-3.1-70B}    & \cellcolor{lightgreen!25}\textbf{45.8} & \cellcolor{lightgreen!25}\textbf{56.0} & \cellcolor{lightgreen!25}\underline{58.4}  &\cellcolor{lightgreen!25}\textbf{50.1}\\
     \multirow{1}{*}{Llama-3.1-405B}    & \cellcolor{lightyellow!25}34.4 & \cellcolor{lightyellow!25}41.7 & \cellcolor{lightyellow!25}48.7  &\cellcolor{lightyellow!25}\underline{40.6}\\
     \multirow{1}{*}{Mistral-0.2-7B}    & \cellcolor{lightred!25}7.0 & \cellcolor{lightred!25}9.5 & \cellcolor{lightred!25}21.0 &\cellcolor{lightred!25}6.9 \\
     \hline
     \hline
     Human        & 92.6 & - & - &-\\
     
     \bottomrule
\end{tabular}
}
\caption{Performance of all LLMs and Humans on StrucText-Eval-Hard.
\textbf{Bolded} text represent the best performance in the column.
\underline{Underlined} text represent the second best performance in the column.}
\label{tab:hard_performance}
\vspace{-5mm}
\end{table}

\begin{figure}[t]
    \centering
    \includegraphics[width=\columnwidth]{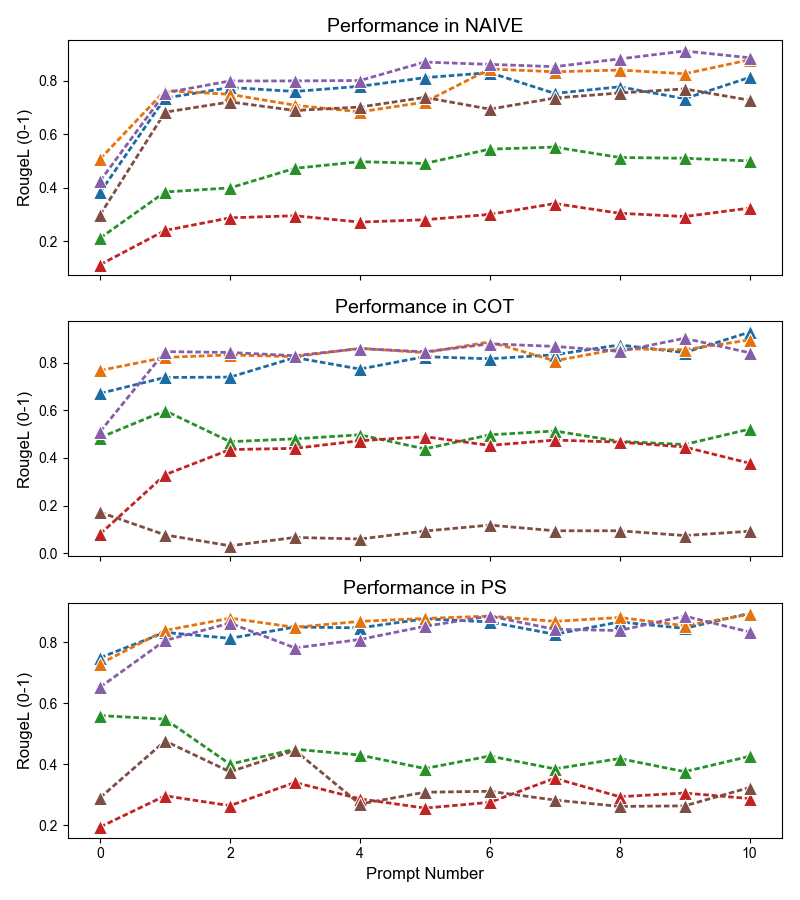}
    \includegraphics[width=0.7\columnwidth]{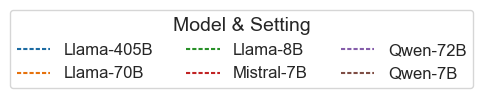}
    \caption{The model's performance on StrucText-Eval-Test under different Few-Shot Demonstration settings.}
    \label{fig:few_shot}
\vspace{-5mm}
\end{figure}

\begin{figure*}[!ht]
\centering
\includegraphics[width=0.24\textwidth]{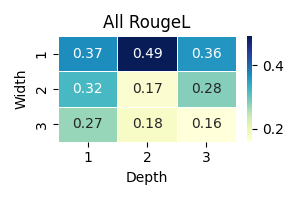}\hfill
\includegraphics[width=0.24\textwidth]{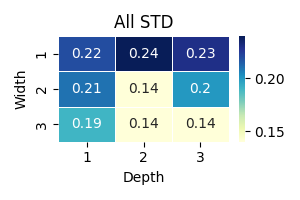}\hfill
\includegraphics[width=0.24\textwidth]{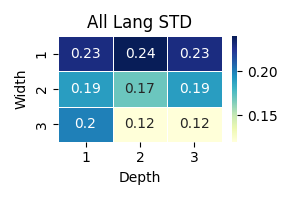}\hfill
\includegraphics[width=0.24\textwidth]{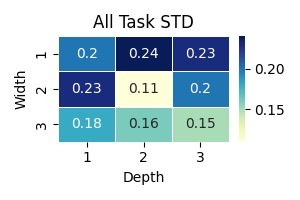}

\includegraphics[width=0.24\textwidth]{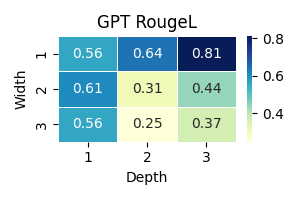}\hfill
\includegraphics[width=0.24\textwidth]{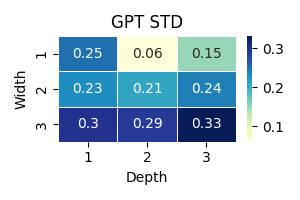}\hfill
\includegraphics[width=0.24\textwidth]{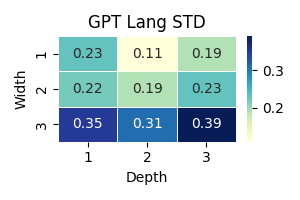}\hfill
\includegraphics[width=0.24\textwidth]{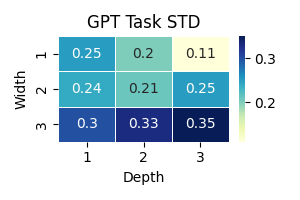}
\caption{Heatmaps illustrating the correlation of RougeL scores and standard deviations (STD) across different models and evaluation criteria. The rows represent different levels of depth, and the columns represent varying levels of width, indicating increasing task complexity. ``All'' refers to combined results across languages and tasks, while ``GPT'' shows results specific to GPT-based models. ``Lang STD'' and ``Task STD'' indicate the variability in performance across different languages and tasks, respectively.}
\label{fig:cor_model}
\vspace{-5mm}
\end{figure*}

\begin{figure*}[!ht]
    \centering
    \resizebox{0.95\textwidth}{!}{
    \includegraphics[width=0.49\linewidth]{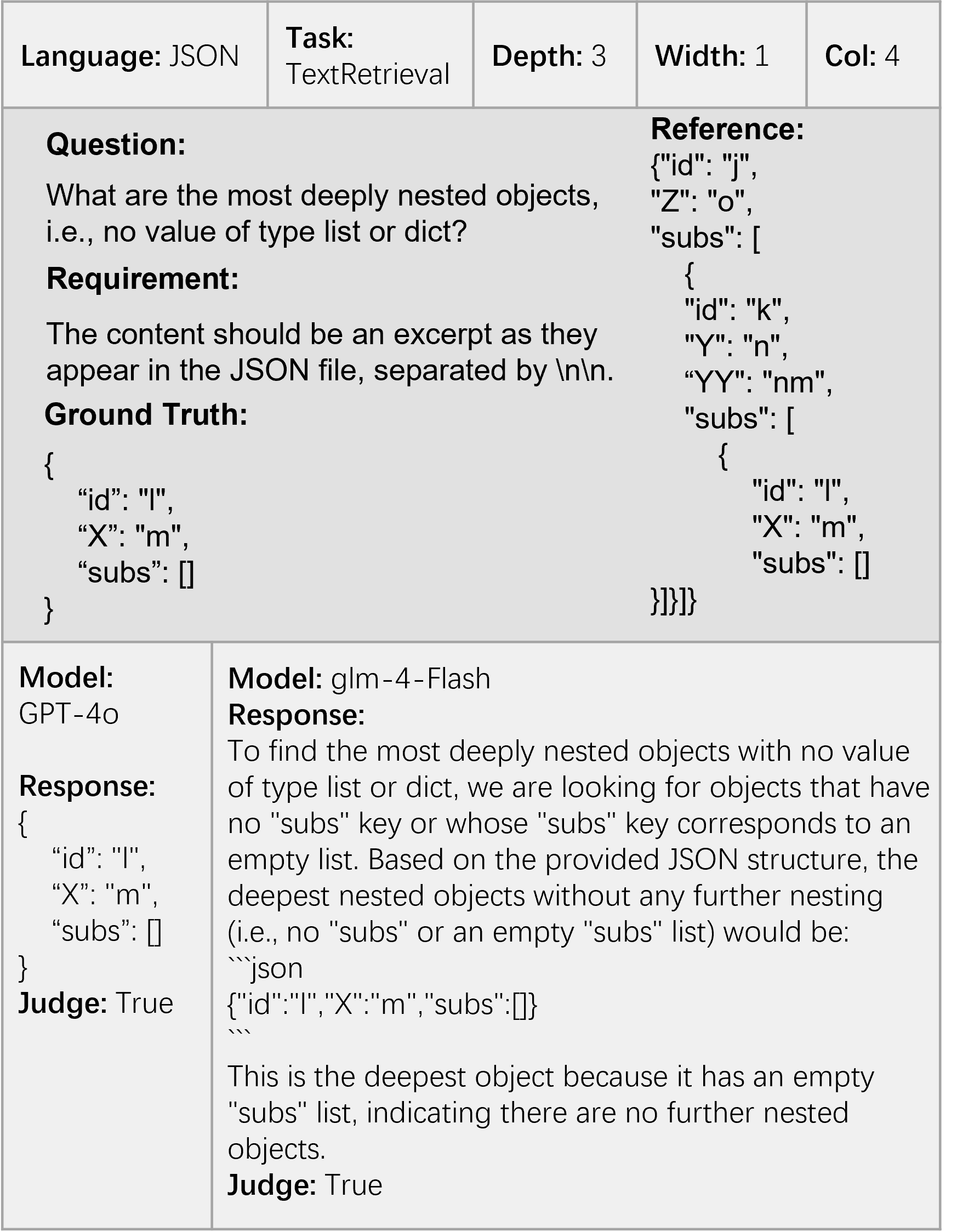}\hfill
    \includegraphics[width=0.49\linewidth]{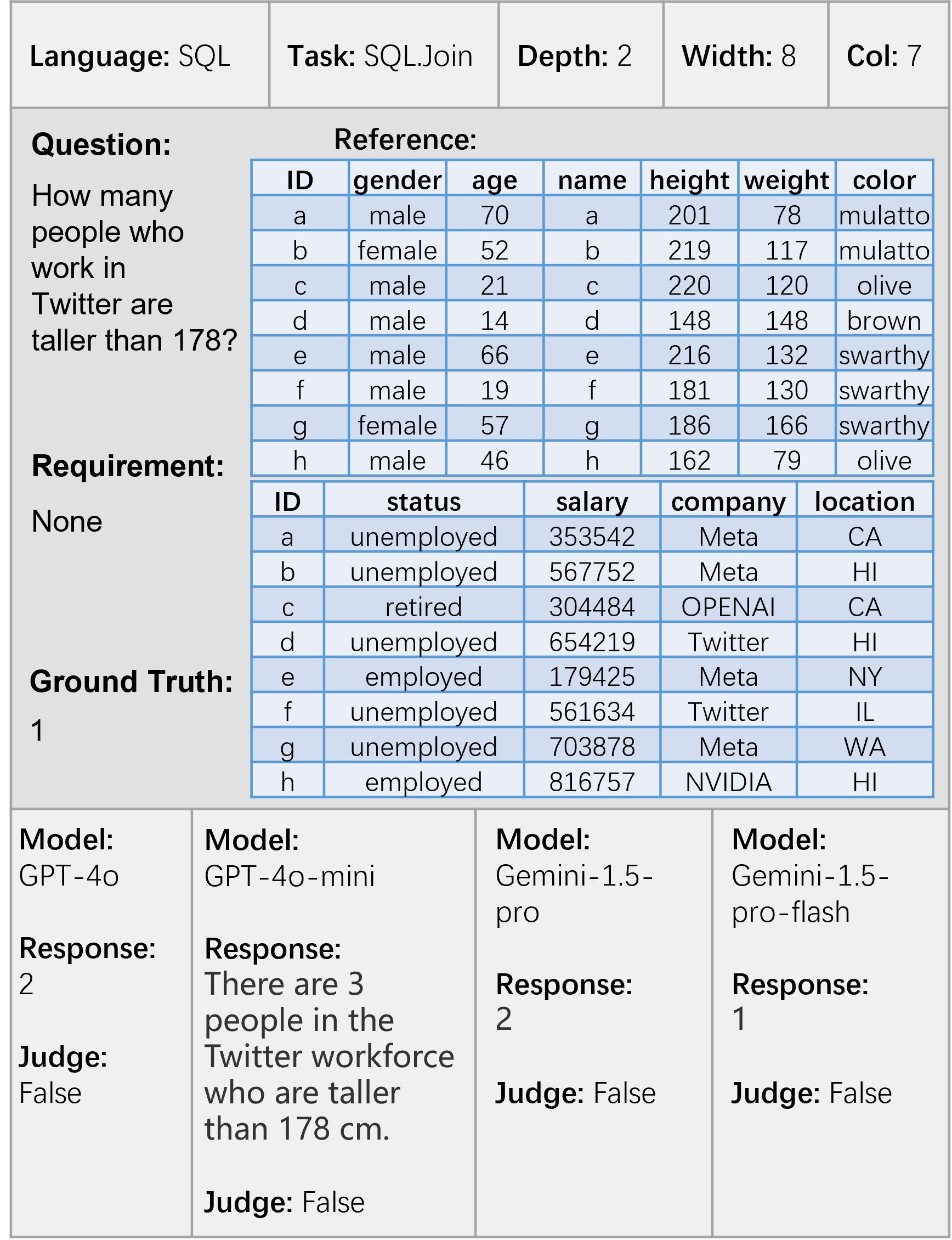}
    }
    \caption{Cases for performance of different LLMs and finetuned stages on Structured Text.}
    \label{fig:case}
    \vspace{-5mm}
\end{figure*}

\subsection{Overall Performance in StrucText-Eval}

The overall performance in StrucText-Eval is presented in Table~\ref{tab:api_overall}, revealing significant variations in the performance of different models across various languages and tasks. For instance, the Qwen2-72B-Instruct model demonstrates optimal performance on JSON-formatted tasks with an 85.8\% accuracy under the ``Naive'' prompt. It also achieves notable results in YAML and CSV tasks, with accuracies of 82.7\% and 86.4\%, respectively. In contrast, the Meta-Llama-3.1-8B-Instruct-Turbo model performs poorly under the same settings, achieving only 64.6\% accuracy on LaTeX tasks. 
Manually injected hints (w/ hint) generally improve model performance, particularly in tasks requiring deep reasoning, such as those involving YAML and JSON.
For example, the Meta-Llama-3.1-70B-Instruct-Turbo model’s accuracy improves from 75.4\% under the ``Naive'' prompt to 84.9\% with the ``w/ Hint'' strategy. 
However, with ``Self-CoT'' and ``PS-CoT'' prompts, specific models like Qwen2-7B-Instruct exhibit lower accuracy across multiple tasks, especially when handling complex structures such as XML and Tree data, performing significantly worse compared to other prompting methods.

These performance disparities can be primarily attributed to training sample biases and the influence of different prompting strategies. JSON, being a widely used format in internet data, is frequently encountered by many large models during training, leading to a pronounced advantage in handling JSON-formatted tasks—a clear manifestation of training sample bias. Moreover, the choice of prompting strategy directly affects a model's inference capabilities. The ``w/ Hint'' method, which introduces human reasoning rules, compensates for the model's limitations in reasoning through complex structures. Conversely, while the ``Self-CoT'' and ``PS-CoT'' approaches encourage step-by-step reasoning, they often result in logical inconsistencies and reasoning errors in complex tasks due to the requirement for autonomous generation of reasoning paths.

\subsection{Overall Performance on StrucText-Eval Hard}
Table \ref{tab:hard_performance} presents the performance of various models on the StrucText-Eval Hard dataset, characterized by more complex tasks with longer sequences and deeper structures. This complexity results in a significant performance decline across all models. For instance, the accuracy of the Qwen2-72B-Instruct model decreases from 78.4\% to 65.0\%, while the Meta-Llama-3.1-70B-Instruct-Turbo model's accuracy drops sharply from 75.4\% to 43.2\%. Unlike the standard dataset, the Hard dataset demands more advanced reasoning skills, and even with the ``w/ Hint'' strategy, models achieve only limited improvements, in contrast to the substantial gains observed in more straightforward contexts. Notably, human accuracy on StrucText-Eval-Hard reaches 95.7\%, significantly surpassing that of the best-performing large language models (LLMs), highlighting a considerable gap in models' capabilities for structured reasoning.

This performance gap can be primarily attributed to biases in training data and the limitations of current prompting methods. The StrucText-Eval Hard dataset, with increased question complexity and depth, requires models to possess enhanced abstraction abilities and a deeper understanding of complex structures. However, most models are trained on relatively more straightforward structured text, which makes them less effective when tackling deeply nested reasoning tasks. Additionally, prompting methods like ``w/ Hint'' fail to achieve human-level understanding in multi-layered scenarios. The differences in prompting methods become more pronounced with increased complexity; more straightforward methods, such as Self-CoT, need to be revised for guiding models through multi-step reasoning in these challenging contexts. While the ``3-shot demonstration'' approach significantly improves model performance, the simpler ``simple 3-shot'' method, despite following similar reasoning rules, fails to match the former due to its insufficient complexity.

\subsection{Few-Shot Demonstration on Structural Text Inference}

Figure \ref{fig:few_shot} demonstrates that model performance improves with an increasing number of demonstrations under Few-Shot settings. In the 3-shot scenario, GPT-4 achieves an accuracy of 69.5\%, significantly outperforming models like Gemini-Pro-Flash and Mistral, which remain around 21\% or lower. The Qwen-2-72B-Instruct model shows steady improvement as more examples are provided, although it continues to trail behind GPT-4. Generally, performance increases from 1-shot to 3-shot, but the gains become less pronounced at 5-shot, with some models showing overfitting. In contrast, the performance of CoT and PS approaches remains less consistent as the number of demonstrations increases.

This trend suggests that a more significant number of examples helps models to understand problem structures and reasoning processes better, thereby enhancing their inference capabilities. However, providing too many examples can lead to models overfitting to specific patterns, which diminishes their ability to generalize to new tasks. The quality and diversity of examples are critical—high-quality examples can guide practical reasoning, while poor examples may mislead the models. While few-shot learning enhances model adaptability, those with limited pretraining data or lower parameter counts may struggle to capitalize on this approach entirely. For CoT and PS methods, the reasoning process requires additional steps, which means that simply increasing the number of few-shot demonstrations does not consistently yield performance improvements.

\subsection{Model Performance Across Different Difficulty Levels, Languages, and Tasks}
Figure \ref{fig:cor_model} illustrates the performance variations of models across different languages and tasks.
The two figures on the left reveal that, while numerical differences exist among models, including GPT models, they exhibit a consistent trend: 
Increasing the reference's depth and width results in a significant decline in performance. 
Notably, all models show a high variance in performance when the depth and width are limited, suggesting that the StrucText-Eval Test suite effectively distinguishes the capabilities of most models under these conditions. 

However, for GPT models, substantial variance in performance is observed only when the depth and width increase significantly, indicating that the StrucText-Eval-Hard Test suite is necessary to better differentiate the performance of more advanced models.
Additionally, there is considerable variance in model performance across different languages and tasks, suggesting substantial differences in models' proficiency in handling various linguistic and task-specific challenges. 
This discrepancy is likely due to biases in training samples and the varying difficulty levels of those samples, as suggested by earlier analyses.

\subsection{Correlation Between Different Metrics}
Figure \ref{fig:cor_metrics} presents the correlations between various evaluation metrics.
The high correlation between Human Judge and GPT-4o Judge (0.9937) indicates a strong alignment between GPT-4o’s automated assessments and human evaluations.
Although Exact Match exhibits a notable correlation with Human Judge (0.9501), its stringent criteria often result in scores significantly lower than those of human evaluators, making it less suitable for capturing the diversity and naturalness of model outputs.
Among the metrics, RougeL stands out with a correlation of 0.9932 with Human Judge, demonstrating its effectiveness in capturing surface-level textual similarity while maintaining high consistency with human judgments.
Compared to the more rigid Exact Match and the relatively lower correlation of BLEU, RougeL offers a better balance between textual similarity and evaluation accuracy.

\subsection{Case Study}
Two case studies illustrate the evaluation setup of StrucText-Eval (Figure \ref{fig:case}). In the JSON-based Text Retrieval task, GPT4-Turbo accurately identified deeply nested objects and adhered to the free-text format for outputting dictionary types, reflecting its firm grasp of structured text. Minimax also produced a correct answer but deviated from the prescribed format, a common issue explored in existing research. In contrast, GPT4-Turbo initially failed to merge two tables and deduce the correct record count without fine-tuning in the SQL-based Join task. However, a finetuned model steadily improved, achieving the correct solution after 5100 training steps. This progression demonstrates the importance of task-specific fine-tuning in enhancing models' capabilities in handling complex SQL queries and database structures.


\section{Conclusion} 
\label{sec:conc}
The capability to directly interpret structural-rich text in a free-text format is an essential skill all LLMs require.
In response, we have developed StrucText-Eval to evaluate this capability of LLMs. 
Our findings indicate that the proficiency of current LLMs in training on these structural-rich texts varies depending on user frequency, leading to markedly different outcomes when the same tasks are performed in various languages.
LLMs' understanding of structural-rich texts remains superficially tied to the training data, and these models need a profound understanding of the structure itself. 
This deficiency becomes evident when LLMs encounter complex structures composed of common languages or need to parse structural-rich text by custom languages, resulting in significant performance degradation.
\section{Limitation}
\label{sec:limitation}


This paper focuses on evaluating LLM's reasoning abilities on structure-rich text by designing a benchmark called StrucText-Eval.
However, StrucText-Eval includes only eight types of structured languages and encompasses a total of 29 different tasks.
Given the vast array of actual structured languages and the myriad methodologies employed beyond these 29 types, StrucText-Eval can only partially represent the LLMs' capacity to understand structure-rich text.
Additionally, due to regional restrictions, we are unable to utilize some highly effective baseline LLMs, such as Gemini and Claude.
Therefore, the conclusions drawn in this paper are based on the assumption that GPT-4 and GPT-4 Turbo represent the top-tier LLMs now.
\section{Ethical Concern}
\label{sec:ethical}


We contend that this article is devoid of ethical concerns for several reasons:
\begin{enumerate}
    \item \textbf{Nature of StrucText-Eval Content}: StrucText-Eval is primarily composed of structured language syntax and some nonsensical strings, which do not present potential ethical issues such as gender bias or racial discrimination.
    \item \textbf{Objective Presentation of Experimental Results}: The experimental results pertaining to StrucText-Eval objectively demonstrate the comprehension abilities of various large models on structure-rich text included in the benchmark. We have thoroughly validated the outputs and assessment details of the models to ensure that the entire evaluation adheres to the experimental setup and maintains objectivity.
    \item \textbf{Completion of Manual Tasks}: All manual tasks associated with this study were conducted by the authors themselves, thereby eliminating any issues of unfair labor practices or unethical cost imposition.
\end{enumerate}

\bibliography{acl}


\appendix

\section{Detail about Manual Works}
\label{ap:manual}

This paper involves the manual works in writing Question Templates and the acquisition of human performance on StrucText-Eval-Hard.
All manual works are carried out by the authors of this paper, so there is no payment for the works.
The three authors collectively completed the writing and verification of 29 Question Templates, and all templates along with the dataset have been made publicly available on the same website.
Moreover, each of the three authors provided responses to 500 identical questions in StrucText-Eval-Hard, with each author dedicating approximately 17 hours.
Thus, the human performance results presented in Table~\ref{tab:hard_performance} are calculated based on the average scores across these 1500 responses.

\section{Other Metrics}
\label{ap:statistic}


Given the substantial expense in evaluating all results using multiple metrics, we selected a subset of 300 test results for each model on the StrucText-Hard dataset, using a naive prompting method for assessment.
The complete evaluation results are presented in Table~\ref{tab:other_metrics}.

\begin{table}[t]
\resizebox{\columnwidth}{!}{

\begin{tabular}{lcccccc} \toprule \textbf{Model} & \textbf{Human} & \textbf{GPT-4o} & \textbf{4o-Mini} & \textbf{RougeL} & \textbf{BLEU} & \textbf{EM} \\ \midrule
GPT-4o-Turbo        & \textbf{56.13} & \textbf{55.75} & \textbf{51.00} & \textbf{51.1} & \underline{45.94} & \textbf{40.31} \\
GPT-4o-Mini         & 36.15 & 36.02 & 40.73 & 39.3 & \textbf{46.08} & 33.93 \\
Gemini1.5-Pro       & 12.39 & 12.80 & 10.62 & 11.2 & 12.60 & 8.75 \\
Gemini1.5-Pro-Flash & 13.83 & 13.19 & 12.96 & 12.9 & 14.01 & 9.67 \\
GLM-4-Plus          & \underline{52.90} & \underline{52.62} & \underline{46.02} & \underline{47.3} & 32.75 & \underline{38.27} \\
GLM-4-Flash         & 41.50 & 41.34 & 38.99 & 40.9 & 37.43 & 34.80 \\
\hline
\hline
QWen-2-7B           & 32.95 & 31.99 & 30.10 & 29.6 & 27.98 & 18.70 \\
QWen-2-72B          & \underline{40.87} & \underline{38.66} & 31.24 & \underline{42.5} & \underline{37.76} & \textbf{35.67} \\
Llama-3.1-8B        & 21.78 & 21.98 & 22.36 & 22.3 & 20.88 & 14.75 \\
Llama-3.1-70B       & \textbf{46.64} & \textbf{41.38} & \textbf{40.83} & \textbf{45.8} & \textbf{41.50} & \underline{27.46} \\
Llama-3.1-405B      & 35.01 & 35.97 & \underline{35.88} & 34.4 & 28.00 & 21.29 \\
Mistral-0.2-7B      & 7.85 & 7.33 & 7.32 & 7.0 & 5.09 & 4.47 \\ 
\bottomrule \end{tabular}

}
\caption{Performance of all LLMs and Humans on StrucText-Eval-Hard based on different metrics~(1,000 samples for each metrics).}
\label{tab:other_metrics}

\end{table}

\section{Detail Prompt}
\label{ap:prompt}


The prompts used in the experiment can be categorized into three types:
Example of Base Prompts are shown in Tab.~\ref{tab:baseprompt}.
Example of CoT Prompts are shown in Tab.~\ref{tab:cotprompt}.
Example of Few-Shot Prompts are shown in Tab.~\ref{tab:fewshot}.
Example of Rule Hints are shown in Tab.~\ref{tab:rulehint}.

\begin{table*}[!ht]
\centering
\resizebox{\textwidth}{!}{
\begin{tabular}{p{\textwidth}}
\toprule
\# -*- coding: utf-8 -*-\\

Variables:\\
!<INPUT 0>! -- Language\\
!<INPUT 1>! -- Question\\
!<INPUT 2>! -- Reference\\
!<INPUT 3>! -- Requirement\\

<commentblockmarker>\#\#\#</commentblockmarker>\\
you are a !<INPUT 0>! file parser, you are required to answer questions pertaining to the given !<INPUT 0>! file.\\
\\
\#\#\# Question:\\
!<INPUT 1>!\\
\\
\#\#\# Reference:\\
!<INPUT 2>!\\
\\
\#\#\# Requirement:\\
!<INPUT 3>!\\
\\
Please follow the format below for your output:\\
\\
\#\#\# Answer:\\
xxxxx\\
\bottomrule
\end{tabular}}
\caption{Prompt of \texttt{Naive Prompt} method}
\label{tab:baseprompt}
\end{table*}

\begin{table*}[!ht]
\centering
\resizebox{\textwidth}{!}{
\begin{tabular}{p{\textwidth}}
\toprule
\# -*- coding: utf-8 -*-\\

Variables:\\
!<INPUT 0>! -- Language\\
!<INPUT 1>! -- Question\\
!<INPUT 2>! -- Reference\\
!<INPUT 3>! -- Requirement\\

<commentblockmarker>\#\#\#</commentblockmarker>\\
you are a !<INPUT 0>! file parser, you are required to answer questions pertaining to the given !<INPUT 0>! file.\\
\\
\#\#\# Question:\\
!<INPUT 1>!\\
\\
\#\#\# Reference:\\
!<INPUT 2>!\\
\\
\#\#\# Requirement:\\
!<INPUT 3>!\\
\\
Please follow the format below for your output:\\
\\
\#\#\# Reasoning Prcess:\\
xxxx\\
\\
\#\#\# Answer:\\
xxxxx\\
\bottomrule
\end{tabular}}
\caption{Prompt of \texttt{CoT} method}
\label{tab:cotprompt}
\end{table*}

\begin{table*}[!ht]
\centering
\resizebox{\textwidth}{!}{
\begin{tabular}{p{\textwidth}}
\toprule
\# -*- coding: utf-8 -*-\\

Variables:\\
!<INPUT 0>! -- Language\\
!<INPUT 1>! -- Demonstration\\
!<INPUT 2>! -- Question\\
!<INPUT 3>! -- Reference\\
!<INPUT 4>! -- Requirement\\

<commentblockmarker>\#\#\#</commentblockmarker>\\
you are a !<INPUT 0>! file parser, you are required to answer questions pertaining to the given !<INPUT 0>! file.\\
\\
\#\#\# Demonstration:\\
!<INPUT 1>!\\
\\
\#\#\# Question:\\
!<INPUT 2>!\\
\\
\#\#\# Reference:\\
!<INPUT 3>!\\
\\
\#\#\# Requirement:\\
!<INPUT 4>!\\
\\
Please follow the format below for your output:\\
\\
\#\#\# Answer:\\
xxxxx\\
\bottomrule
\end{tabular}}
\caption{Prompt of \texttt{Few Shot} method}
\label{tab:fewshot}
\end{table*}

\begin{table*}[!ht]
\centering
\resizebox{\textwidth}{!}{
\begin{tabular}{p{\textwidth}}
\toprule

\# -*- coding: utf-8 -*-\\

Variables:\\
!<INPUT 0>! -- Language\\
!<INPUT 1>! -- Question\\
!<INPUT 2>! -- Reference\\
!<INPUT 3>! -- Requirement\\
!<INPUT 4>! -- Rule Hint\\

<commentblockmarker>\#\#\#</commentblockmarker>\\
you are a !<INPUT 0>! file parser, you are required to answer questions pertaining to the given !<INPUT 0>! file.\\
\\
\#\#\# Question:\\
!<INPUT 1>!\\
\\
\#\#\# Reference:\\
!<INPUT 2>!\\
\\
\#\#\# Requirement:\\
!<INPUT 3>!\\
\\
\#\#\# Rule Hint:\\
!<INPUT 4>!\\
\\
Please follow the format below for your output:\\
\\
\#\#\# Answer:\\
xxxxx\\
\bottomrule
\end{tabular}}
\caption{Prompt of \texttt{\textbackslash w Hint} method}
\label{tab:rulehint}
\end{table*}

\section{Examples for All Languages \& Tasks}
\label{ap:samples}

In this section, we provide detailed examples for each language we discuss, illustrating how specific tasks are executed within those languages. 
These examples are meant to offer clear insights into the application and utility of each language in various contexts. 
Through these demonstrations, readers can better understand the unique features and capabilities of each language when applied to different tasks.

\subsection{Tree}
\begin{figure}[!ht]
\caption{Sample input and tasks of Tree.}
\label{fig:eg_tree}
\begin{center}
   \includegraphics[width=\linewidth]{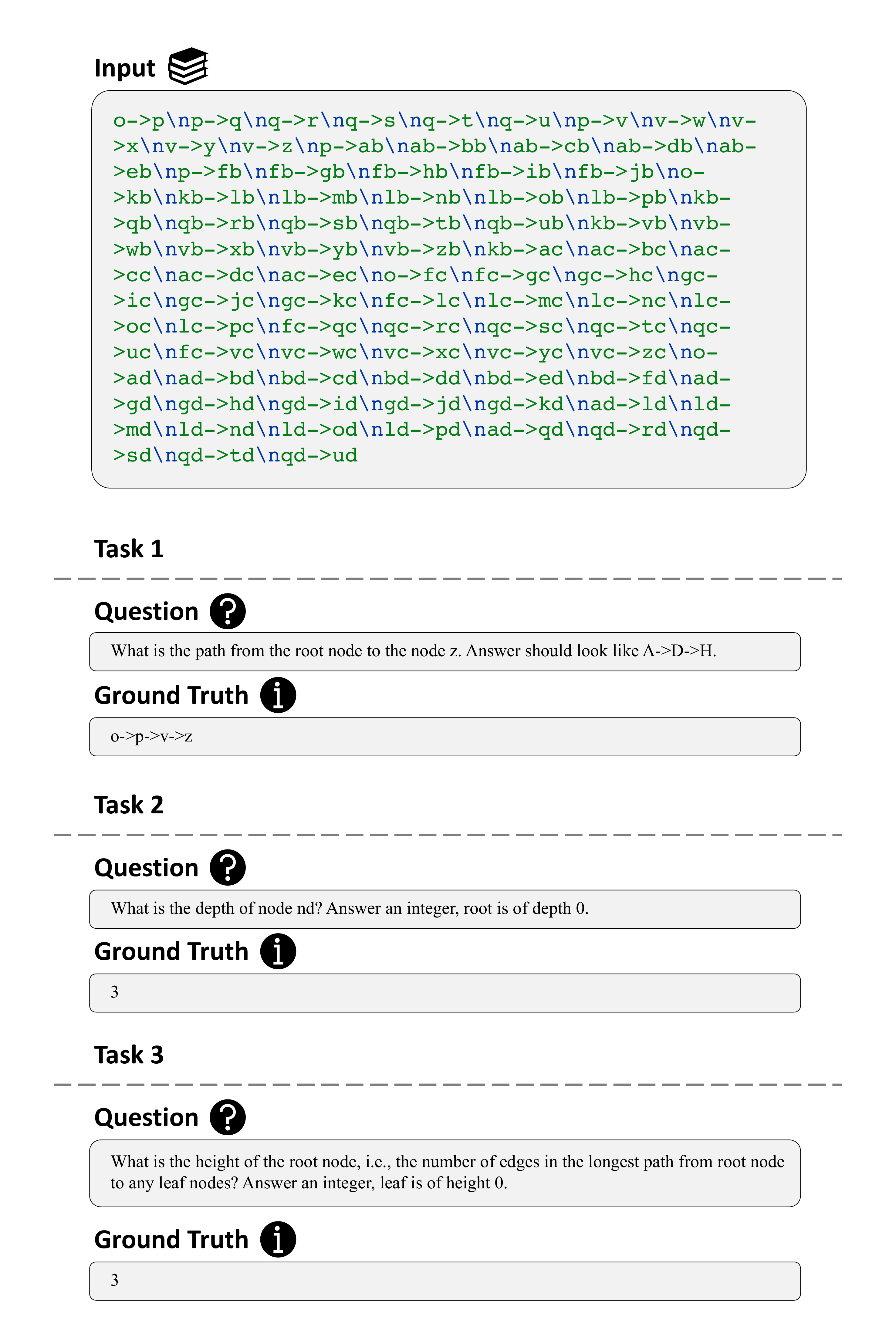}
\end{center}
\end{figure}

See~\autoref{fig:eg_tree}.

\subsection{Tabular}

\begin{figure}[!ht]
\caption{Sample input and tasks of tabular data.}
\label{fig:eg_tab}
\begin{center}
   \includegraphics[width=\linewidth]{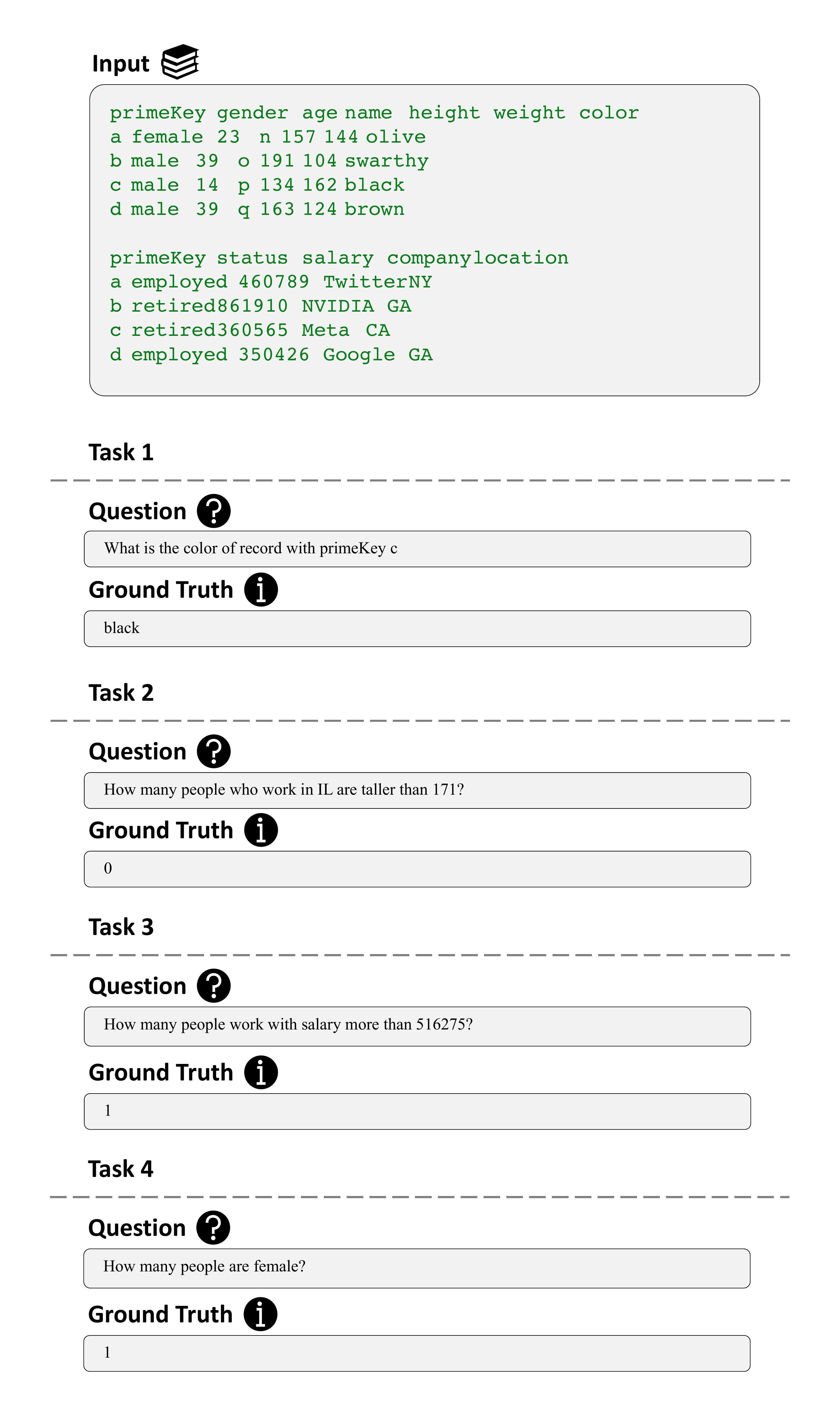}
\end{center}
\end{figure}

See~\autoref{fig:eg_tab}.

\subsection{JSON}

\begin{figure}[!ht]
\caption{Sample input and tasks of JSON.}
\label{fig:eg_json}
\begin{center}
   \includegraphics[width=\linewidth]{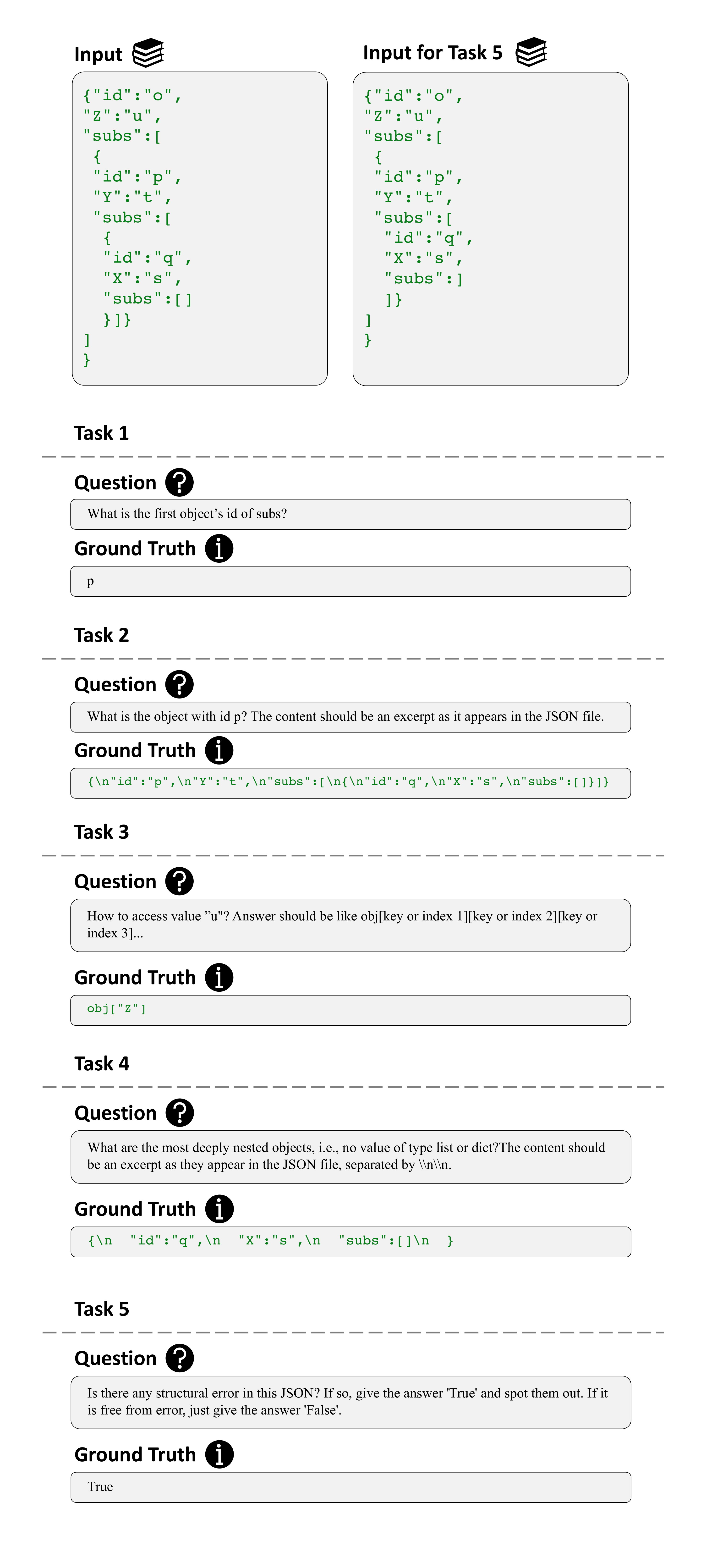}
\end{center}
\end{figure}

See~\autoref{fig:eg_json}.

\subsection{YAML}

\begin{figure}[!ht]
\caption{Sample input and tasks of YAML.}
\label{fig:eg_yaml}
\begin{center}
   \includegraphics[width=\linewidth ]{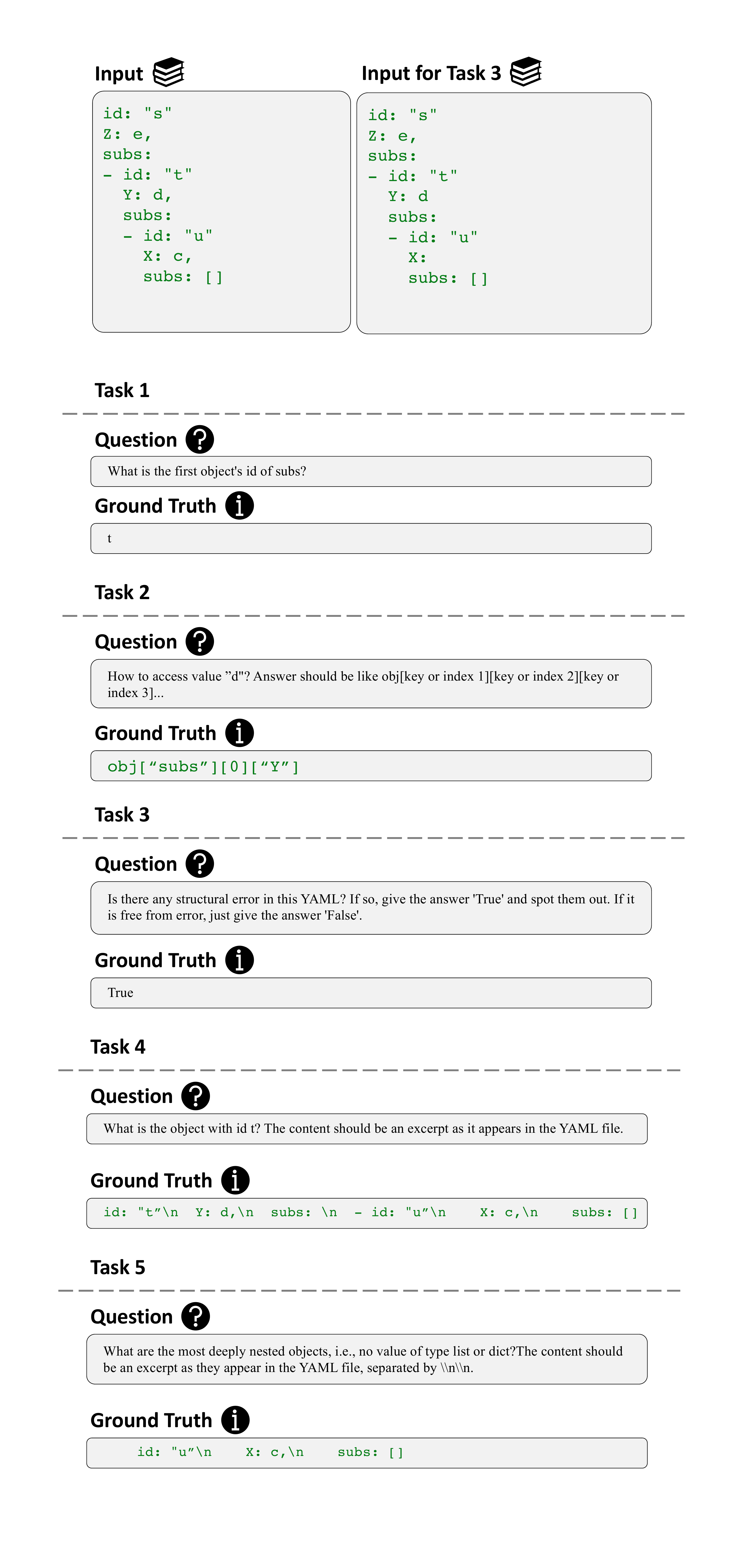}
\end{center}
\end{figure}

See~\autoref{fig:eg_yaml}.

\subsection{XML}

\begin{figure}[!ht]
\caption{Sample input and tasks of XML.}
\label{fig:eg_xml}
\begin{center}
   \includegraphics[width=\linewidth]{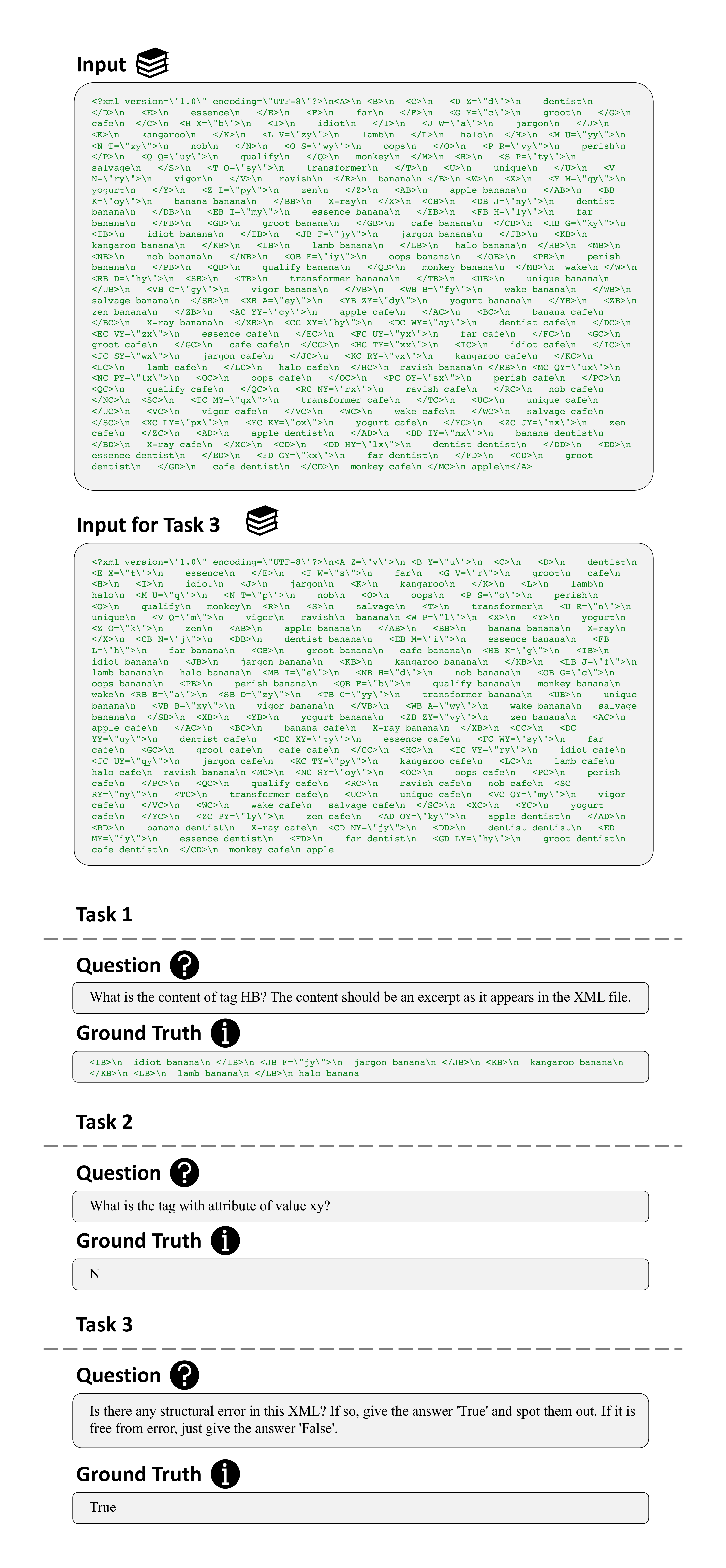}
\end{center}
\end{figure}

See~\autoref{fig:eg_xml}.

\subsection{LaTeX}

\begin{figure}[!ht]
\caption{Sample input and tasks of LaTeX.}
\label{fig:eg_latex}
\begin{center}
   \includegraphics[width=\linewidth]{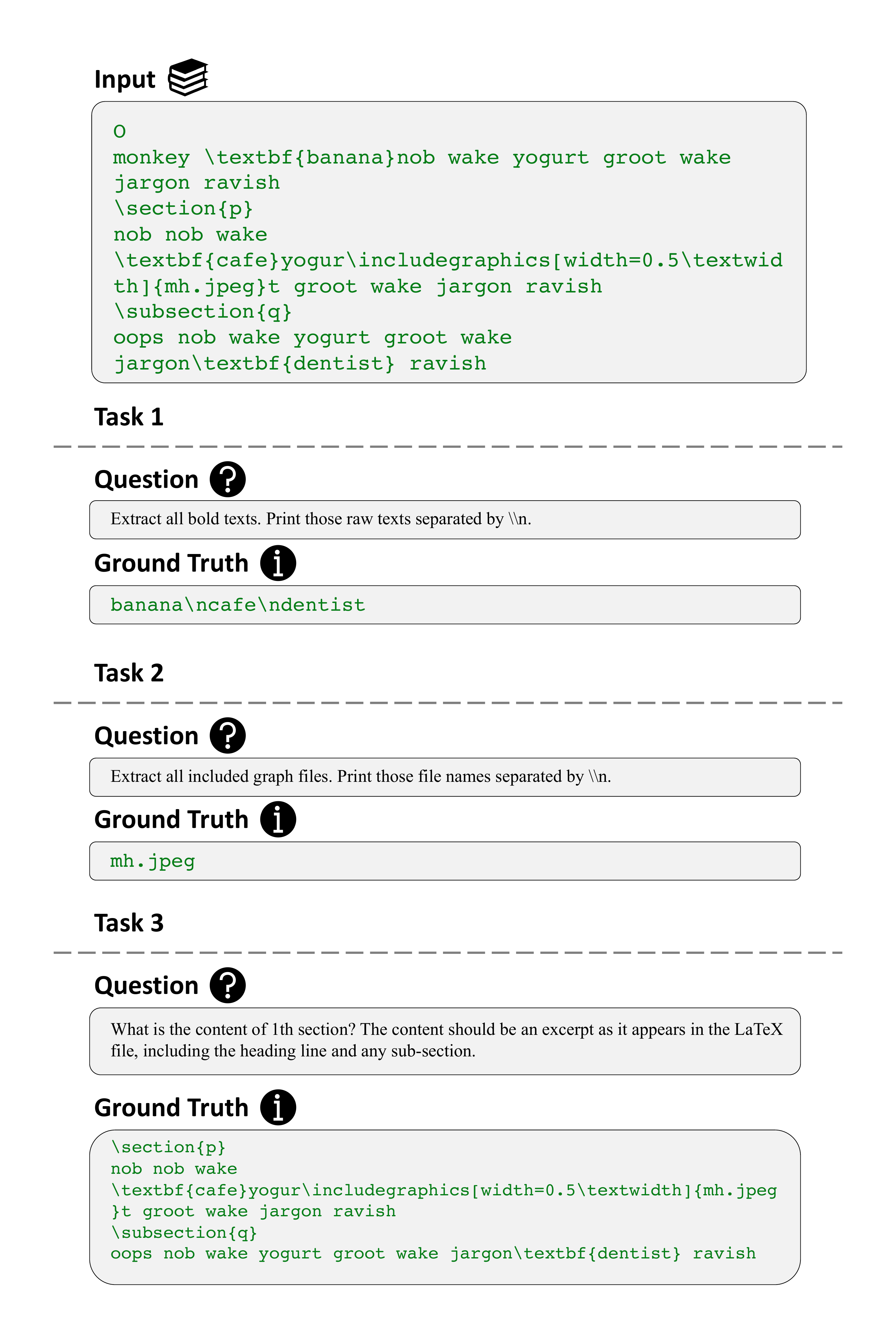}
\end{center}
\end{figure}

See~\autoref{fig:eg_latex}.

\subsection{Markdown}

\begin{figure}[!ht]
\caption{Sample input and tasks of Markdown.}
\label{fig:eg_mkd}
\begin{center}
   \includegraphics[width=\linewidth]{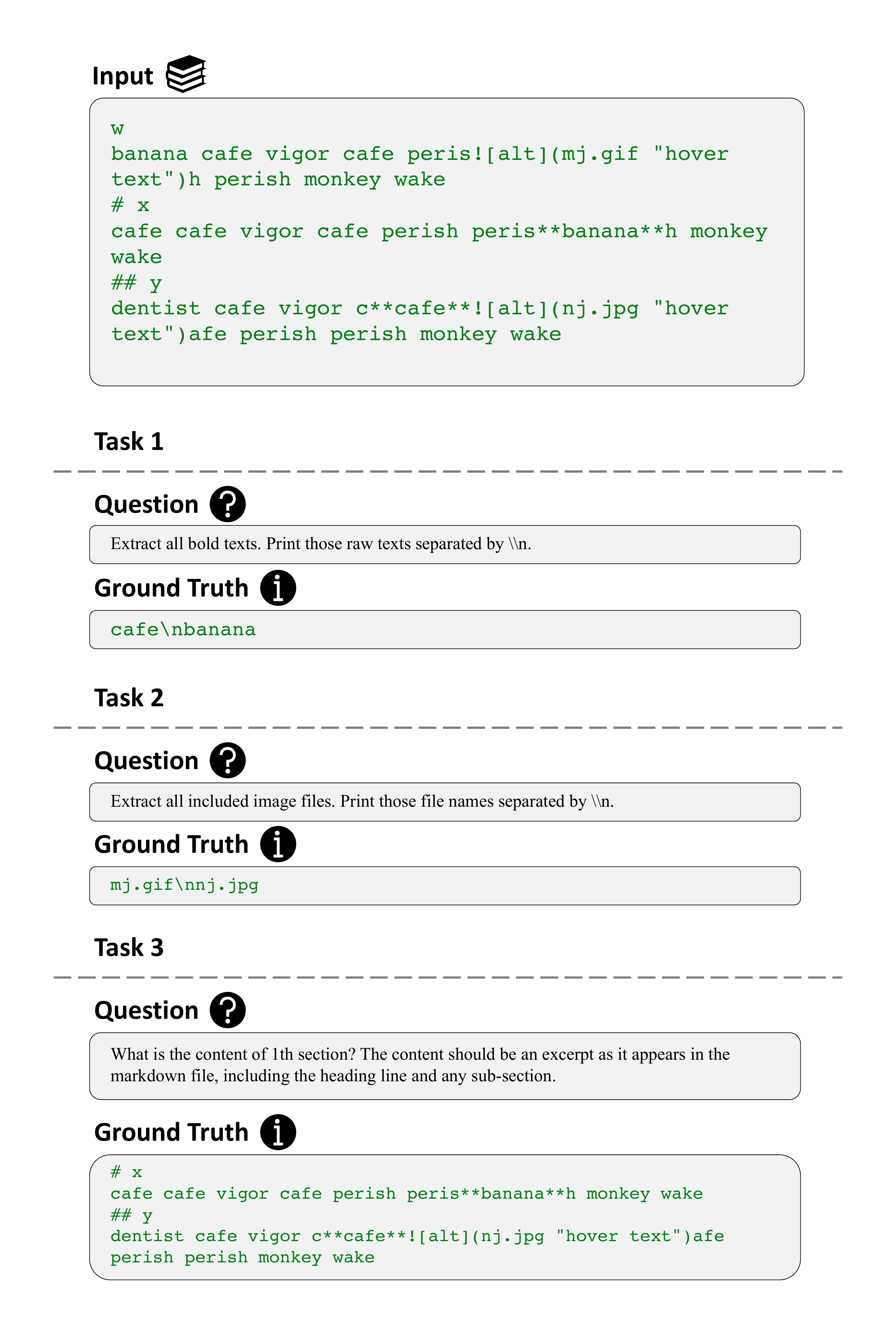}
\end{center}
\end{figure}

See~\autoref{fig:eg_mkd}.

\subsection{Org}

\begin{figure}[!ht]
\caption{Sample input and tasks of Org.}
\label{fig:eg_org}
\begin{center}
   \includegraphics[width=\linewidth]{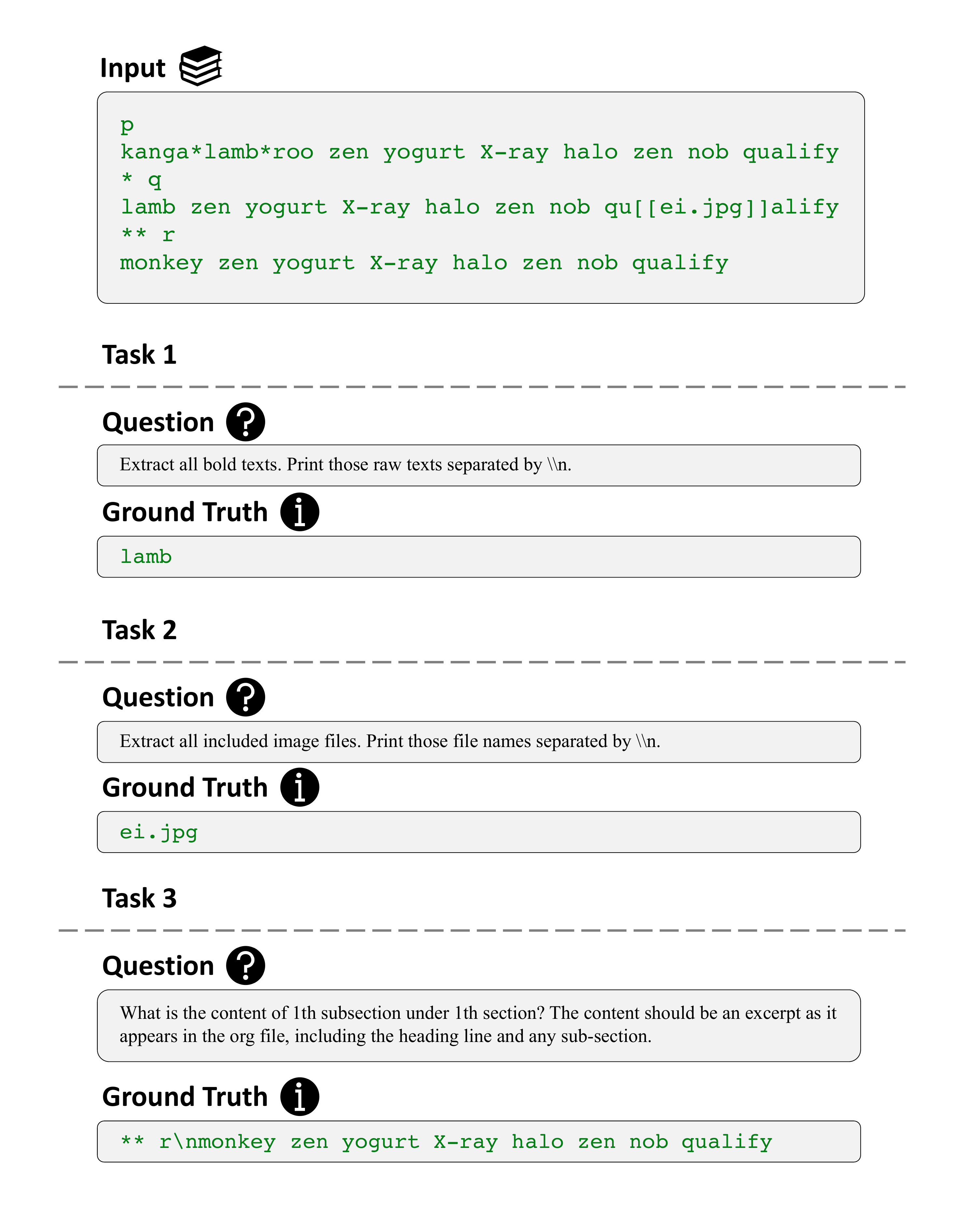}
\end{center}
\end{figure}

See~\autoref{fig:eg_org}.

\section{Rules \& Rule Hints}
\label{ap:rules}
We list all the rules in Regular Express in this section, and list all the hints for these rules in Lis.~\ref{list:rulehint}.

\subsection{Tree}

We build tree-structured input as a list of edges in a tree, in a format of ``\texttt{father->child}’’, separated by newline.

\[
\begin{aligned}
identifier&\coloneqq \texttt{[a-z]+}\\
Edge&\coloneqq identifier\texttt{->}identifier\\
Tree&\coloneqq Edge(\texttt{\textbackslash n}Edge)\texttt{*}\\
InputFile&\coloneqq Tree\\
\end{aligned}
\]

\subsection{Tabular}
Formally, input texts are classified as tabular data given that they are composed of a list of newline separated lines, each of which is a list of text cells delimited by comma.

\[
\begin{aligned}
head &\coloneqq \texttt{[A-Z][a-z]*}\\
cell &\coloneqq \texttt{[A-Za-z0-9]+}\\
headline &\coloneqq identifier(, identifier)\texttt{*}\\
subline &\coloneqq cell(, cell)\texttt{*}\\
Tabular &\coloneqq headline(\texttt{\textbackslash n}subline)\texttt{+}\\
InputFile&\coloneqq Tabular\\
\end{aligned}
\]

\subsection{JSON}\label{ap:json_rule}

Due to the inherit hierarchy structure of Object Notations, we adopted a recursive scheme to define our input texts.

\[
\begin{aligned}
lb_{(left\ bracket)}&\coloneqq \texttt{[[]}\\
rb&\coloneqq \texttt{[]]}\\
val&\coloneqq \texttt{[a-z]+}\\
key&\coloneqq \texttt{[A-Z]+}\\
JSON&\coloneqq \{\\
&\texttt{"id":"}val\texttt{"}\\
&\texttt{"subs":}lbrb|lbJSON(, \texttt{\textbackslash n}JSON\\
&)\texttt{*}rb\\
&\texttt{("}key\texttt{":"}val\texttt{"\textbackslash n)+}\\
&\}\\
InputFile&\coloneqq JSON\\
\end{aligned}
\]

\subsection{YAML} \label{ap:yaml_rule}
The rules for constructing YAML and XML input are similarly recursive.

\[
\begin{aligned}
YAML&\coloneqq \\
&\texttt{id}:val\\
&\texttt{subs}:lbrb|(\texttt{\textbackslash n}(\texttt{\textbackslash t})*\texttt{- }YAML)\\
&+(key:val\texttt{\textbackslash n})\texttt{+}\\
InputFile&\coloneqq YAML\\
\end{aligned}
\]
\subsection{XML} \label{ap:xml_rule}
\[
\begin{aligned}
firstline&\coloneqq \texttt{<?xml version="1.0"}\\
&texttt{encoding=}\text{``UTF-8''}\texttt{?>}\\
XML &\coloneqq\\
&firstline\\
&XMLObject\\
tag&\coloneqq \texttt{[A-Z]+}\\
val&\coloneqq \texttt{[a-z]+}\\
attr&\coloneqq \texttt{[A-Z]+="}val\texttt{"}\\
content&\coloneqq \texttt{[a-z \textbackslash n\textbackslash t]*}\\
XMLObject&\coloneqq\\
&\texttt{<}tag(\texttt{ }attr)*\texttt{>}\\
&((\texttt{\textbackslash t})*XMLObject)*\\
&content\\
&\texttt{</}tag\texttt{>}\\
InputFile&\coloneqq XML\\
\end{aligned}
\]

\subsection{LaTeX} \label{ap:ltx_rule}
In LaTeX input texts, we include \texttt{textbf} and \texttt{includegraphics} commands to accommodate for the text retrieval tasks. The headings serve as anchors for structure traversal.
\[
\begin{aligned}
command &\coloneqq \texttt{\textbackslash}(\texttt{section}|\texttt{subsection}|\\
&\texttt{subsubsection})\\
heading &\coloneqq command\texttt{\{[a-z]+\}}|\texttt{[a-z]+}\\
inclg &\coloneqq\\ &\texttt{\textbackslash includegraphics[width=}\\
&\texttt{0.5\textbackslash textwidth]}\{\texttt{[a-z]+[.]}\\
&\texttt{(png|jpg|jpeg|gif)}\}\\
bf &\coloneqq \texttt{\textbackslash textbf}\{\texttt{[a-z ]+}\}\\
content &\coloneqq (\texttt{[a-z ]}|bf|inclg)\texttt{+}\\
LaTeX &\coloneqq heading\texttt{\textbackslash n}content (\texttt{\textbackslash n}LaTeX)*\\
InputFile&\coloneqq LaTeX\\
\end{aligned}
\]

\subsection{Markdown}
In markdown input texts, the syntax counterparts for heading, text face and including figure are employed in our dataset.

\[
\begin{aligned}
	heading&\coloneqq \texttt{[\#]* [a-z]+}\\
	inclg&\coloneqq \texttt{!}lb\texttt{alt}rb\texttt{\textbackslash([a-z]+[.](png}\\
 &\texttt{|jpg|jpeg|gif)}\\
 &\texttt{"hover text"\textbackslash)}\\
	bf&\coloneqq \texttt{[*]\{2\}[a-z ]+[*]\{2\}}\\
    content &\coloneqq (\texttt{[a-z ]}|bf|inclg)\texttt{+}\\
    Markdown &\coloneqq heading\texttt{\textbackslash n}\\
    &content (\texttt{\textbackslash n}Markdown)*\\
    InputFile&\coloneqq Markdown\\
\end{aligned}
\]

\subsection{Org}
In Org input texts, the syntax is obtained from JSON construction rules by replacing the markups for heading, including figures and bold font face.
\[
\begin{aligned}
	heading&\coloneqq \texttt{[*]* [a-z]+}\\
	inclg&\coloneqq lb\{2\}\texttt{[a-z]+[.](png|jpg|}\\
 &\texttt{jpeg|gif)}rb\{2\}\\
	bf&\coloneqq \texttt{[*][a-z ]+[*]}\\
    content &\coloneqq (\texttt{[a-z ]}|bf|inclg)\texttt{+}\\
    Org &\coloneqq heading\texttt{\textbackslash n}content (\texttt{\textbackslash n}Org)*\\
    InputFile&\coloneqq Org\\
\end{aligned}
\]

\begin{lstlisting}[caption={All rule hints in StrucText-Eval}, label={list:rulehint}]
SQL,Tree,JSON,YAML,XML,Markdown,LaTeX,ORG
To find the value of specific field of record with specified primeKey. You have to first, locate the line with the specific primeKey. Then find the required value under the desired column in that line.
To get the number of people with salary above a threshold, you need to find the table with salary information. Then you go over each line and check the salary field. During the process count only those lines with value of salary strictly greater than the specified threshold towards your final sum. The sum after checking each line is the right answer.
To get the number of female, first find the table with column name ''. Then check each line for field gender, and count these lines with value 'female' towards your final sum. The process applies to finding number of male too.
To get the number of people living in specified city who are also taller than threshold, you need to first join the two table on primeKey, and check each row of joined table for lines that satisfies both condition, i.e., lines with city specified in query and height strictly greater than threshold. The total number of such rows is the right answer.
To answer the height of tree, you need to take a recursive strategy. For each node, you will find its height by first finding its children's heights. Then, the height of node is the maximum subtree heights plus 1. The base case occurs when a node has no children, i.e., it's a leaf node. Leaf's height is defined to be 0, without the need of further queries. Then the height the tree is the height of its root node.
To find the depth of a node, you need to find the number of edges from root to node. You have to start from the root with depth 0 and assign the depth for each node recursively. For any given node, it gets depth of current depth. Increment the depth by 1 before go to its subtree and repeat the process until every node gets a depth.
To get the path from root to a node, you need to find recursively. For any node, you can find the path to the target node by find path from its children to target. Then check each child's output, if any child returns with valid path instead of an empty path indicating target-not-found, the path from node to target is that path from its child to target prepended with itself. The answer can be found by searching with root as starting point.
To find the object with specified id, you need to first parse the json file and get the outermost object, starting from which search the subs field recursively and looking for the desired value in id field for each visited object. Retrieve the content of that object once found.
To find the first object's id of subs, first parse the json file and get the outermost object, in the outermost object's subs list, get the first element. That element is another object, and its id is the answer.
To find the error in the json file, you need to parse the json file and report any syntax error if encountered any. Potential errors include missing ending curly braces.
To get the path to access specified value. You have to do a recursive search along the subs fields, starting from the outermost parsed object. For each visited object, check each fields except for subs, and record the path along the way, i.e., subs inside brackets and index into subs inside brackets, and at which field you find the value.
To get the most deeply nested objects, start from the outermost object, recursively search along the subs fields. For each object, check its subs field, any object with an empty subs is one most deeply nested object.
To find the object with specified id, you need to first parse the yaml file and get the outermost object, starting from which search the subs field recursively and looking for the desired value in id field for each visited object. Retrieve the content of that object once found.
To find the first object's id of subs, first parse the yaml file and get the outermost object, in the outermost object's subs list, get the first element. That element is another object, and its id is the answer.
To find the error in the yaml file, you need to parse the yaml file and report any syntax error if encountered any. Potential errors include missing key before colon.
To get the path to access specified value. You have to do a recursive search along the subs fields, starting from the outermost parsed object. For each visited object, check each fields except for subs, and record the path along the way, i.e., subs inside brackets and index into subs inside brackets, and at which field you find the value.
To get the most deeply nested objects, start from the outermost object, recursively search along the subs fields. For each object, check its subs field, any object with an empty subs is one most deeply nested object.
To find the content of a specific tag, you need to search for desired tag throughout the xml file. Once located, find the surrounding left and right angle, these area is tha starting tag. Then find the ending tag, which is the tag surrounded by angle with exception that right angle is preceded by a slash. The content between starting and ending tags is the answer.
To find the tag name of particular attribute value, just search the file for that value and find the surrounding left and right angles, i.e., boundary of tag. The word next to left angle is tag name.
To find the error in the xml file, you need to parse the xml file and report any syntax error if encountered any. Potential errors include missing ending tags.
To find the bold texts, search for double stars, i.e., **, the content between two occurrences of double stars is the bold texts. Note that the bold range should start from the double stars occurring at i-th spot throughout the whole input file, where i is odd, and end with double stars occurring at jth spot where j is even. For example, text between double stars appearing first and second time.
To find the content of certain section, starting from the headings start with one hashtag, and go to the ith heading as specified in number of sections. Then start from that line, look for j-th heading with 2 hashtags as specified in subsection number. For kth subsubsection, look for kth heading with 3 hashtags starting from the located subsubsection. Stop searching early if the subsection or subsubsection is not queried.
To find the image files, look for texts matching ![*](TARGET "*"), the TARGET part is filename. Star means any text is possible.
To find the bold texts, search for macro textbf, and everything after \\textbf{ and before the first } encountered is bold text.
Note that section title is enclosed by \\section{}, and \\subsection for subsection, \\subsubsection for subsubsection. To find the content of certain section, look for ith section as specified, and start from there look for jth subsection. And from located subsection, look for kth subsubsection as queried. Search may stop early if subsection or subsubsection is not queried.
To find the image files imported, search for pattern \\includegraphics[*]{TARGET}, the TARGET part is the filename. Star means any text is possible.
To find the bold texts, search for single star, i.e., *, the content between two occurrences of single star is the bold texts. Note that the bold range should start from the single star occurring at i-th spot throughout the whole input file, where i is odd, and end with single star occurring at jth spot where j is even. For example, text between single star appearing first and second time.
Note that section, subsection, subsubsection titles are preceded by *, **, *** respectively, with one or more whitespaces in between.  To find the content of certain section, look for ith section as specified, and start from there look for jth subsection. And from located subsection, look for kth subsubsection as queried. Search may stop early if subsection or subsubsection is not queried.
To find the image files, look for texts matching [[TARGET]], the TARGET part is filename
\end{lstlisting}


\section{Detail Setting}
\label{ap:setting}

All experiments and training process is carried out on a three 3090 GPUs service.
The setting of API calling is illustrated in Tab.~\ref{tab:hyperparameter}

\begin{table*}[!ht]
        \centering
        \resizebox{\textwidth}{!}{
        \begin{tabular}{|c|c|c|c|c|}
        \toprule
        \multicolumn{5}{|c|}{Random Seed} \\
        \hline
        torch.manual\_seed & torch.cuda.manual\_seed\_all & numpy.random.seed & random.seed & torch.backends.cudnn.deterministirc \\
        42 & 42 & 42 & 42 & True \\
        \hline
        \hline
        \multicolumn{5}{|c|}{AutoCausalLM} \\
        \hline
        temperature & top\_p & top\_k & num\_beams & max\_new\_token \\
        0.95 & 0.95 & 5 & 2 & 1  \\
        \bottomrule
        \end{tabular}
        }
    \caption{All the parameter setting in our experiments.}
    \label{tab:hyperparameter}
\end{table*}




\end{document}